\documentclass[10pt,twocolumn,letterpaper]{article}

\usepackage[pagenumbers]{wacv} 

\usepackage{adjustbox}
\usepackage[table]{xcolor}
\usepackage{tcolorbox}
\usepackage{threeparttable}
\usepackage{appendix}
\usepackage{placeins}

\providecommand{\sg}{\operatorname{sg}}      
\providecommand{\ind}{\mathbf{1}}            

\definecolor{lgreen}{HTML}{ddf2d1}
\colorlet{lred}{red!15}

\usepackage{xspace}
\newcommand{\SSMM}{S\textsuperscript{2}M\textsuperscript{2}\xspace}

\definecolor{wacvblue}{rgb}{0.21,0.49,0.74}
\usepackage[pagebackref,breaklinks,colorlinks,allcolors=wacvblue]{hyperref}

\def\wacvPaperID{1193} 
\def\confName{WACV}
\def\confYear{2027}

\title{Left-Right-Consistent Iterative Stereo Matching via Cross-Injection}

\author{
  Manuel Knott \quad
  Diogo Luvizon \quad
  Daniel Scharstein \quad
  Elias Vansteenkiste \\
  Meta Reality Labs 
}

\begin{document}
\maketitle
\begingroup \renewcommand{\thefootnote}{} \footnotetext{\noindent GitHub: \href{https://github.com/facebookresearch/stereo_crossinjection}{{https://github.com/facebookresearch/stereo\_crossinjection}}} \endgroup \setcounter{footnote}{0}
\begin{abstract}
State-of-the-art stereo matching models predominantly predict the disparity of only the left (reference) view. However, applications such as per-eye AR/VR rendering, novel view synthesis, or two-view depth fusion require disparity for both views, and need those two maps to agree. The usual workaround of running the network a second time on the horizontally flipped and swapped pair gives two independent predictions that can be inconsistent, especially on transparent surfaces, in textureless regions, and at occlusion boundaries.
We introduce cross-injection (CI), a lightweight extension for iterative (refinement-based) models that can be added to any pretrained left-only iterative matcher. It couples the two passes by injecting the explicit, signed left-right inconsistency as an additional input cue at every step of the refinement stage.
We empirically show that CI substantially reduces left-right disagreement on every benchmark (more than halving it on Booster for FoundationStereo), while largely preserving sharp object boundaries. Furthermore, using two state-of-the-art stereo matching models, we show that CI also reduces the single-view disparity error, making it a worthwhile extension even if only left disparity is needed.
\end{abstract}

\section{Introduction}
\label{sec:intro}

A growing number of vision systems need geometrically consistent disparity for \emph{both} cameras of a stereo rig, not just for a reference view.
Per-eye passthrough rendering on AR/VR headsets reprojects and composites each eye with its own depth~\cite{chaurasia2020passthrough}, so any left-right disagreement introduces noticeable artifacts~\cite{elchemaly2025geometrypassthrough}.
Stereoscopic content must also keep disparity within a viewing-comfort zone~\cite{kim2021visualcomfortrl}. Novel view synthesis from a stereo pair renders new viewpoints from the two input views~\cite{zhou2018stereomag}, so where the two disparity maps disagree the composite tears and ghosts at occlusion boundaries. Two-view depth fusion likewise yields doubled geometry wherever the two estimates disagree. These examples demonstrate the need not just for two accurate depth maps, but also \emph{consistent} ones.

\begin{figure}
    \centering  \includegraphics[width=0.49\linewidth]{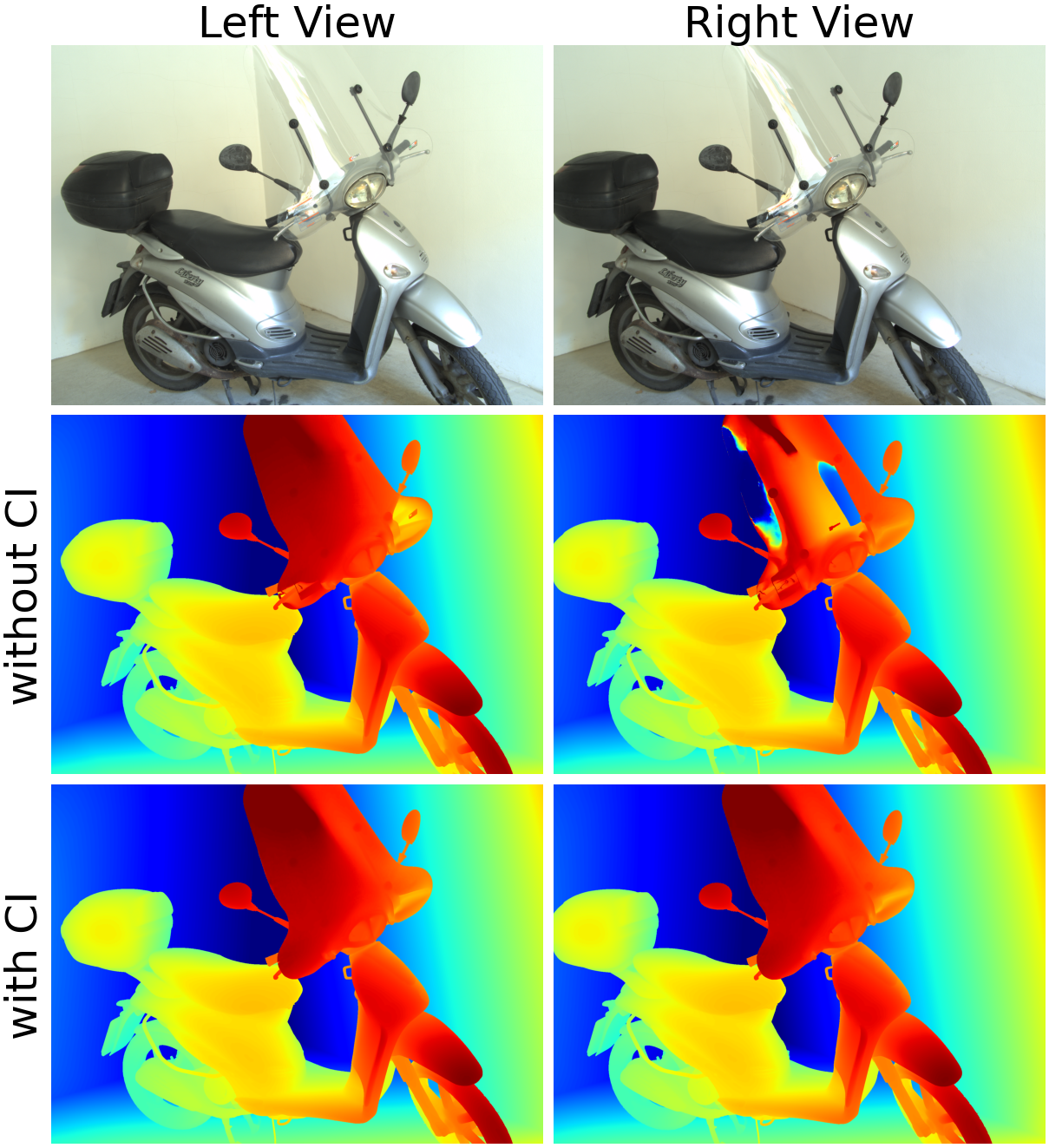}
     \hfill   \includegraphics[width=0.49\linewidth]{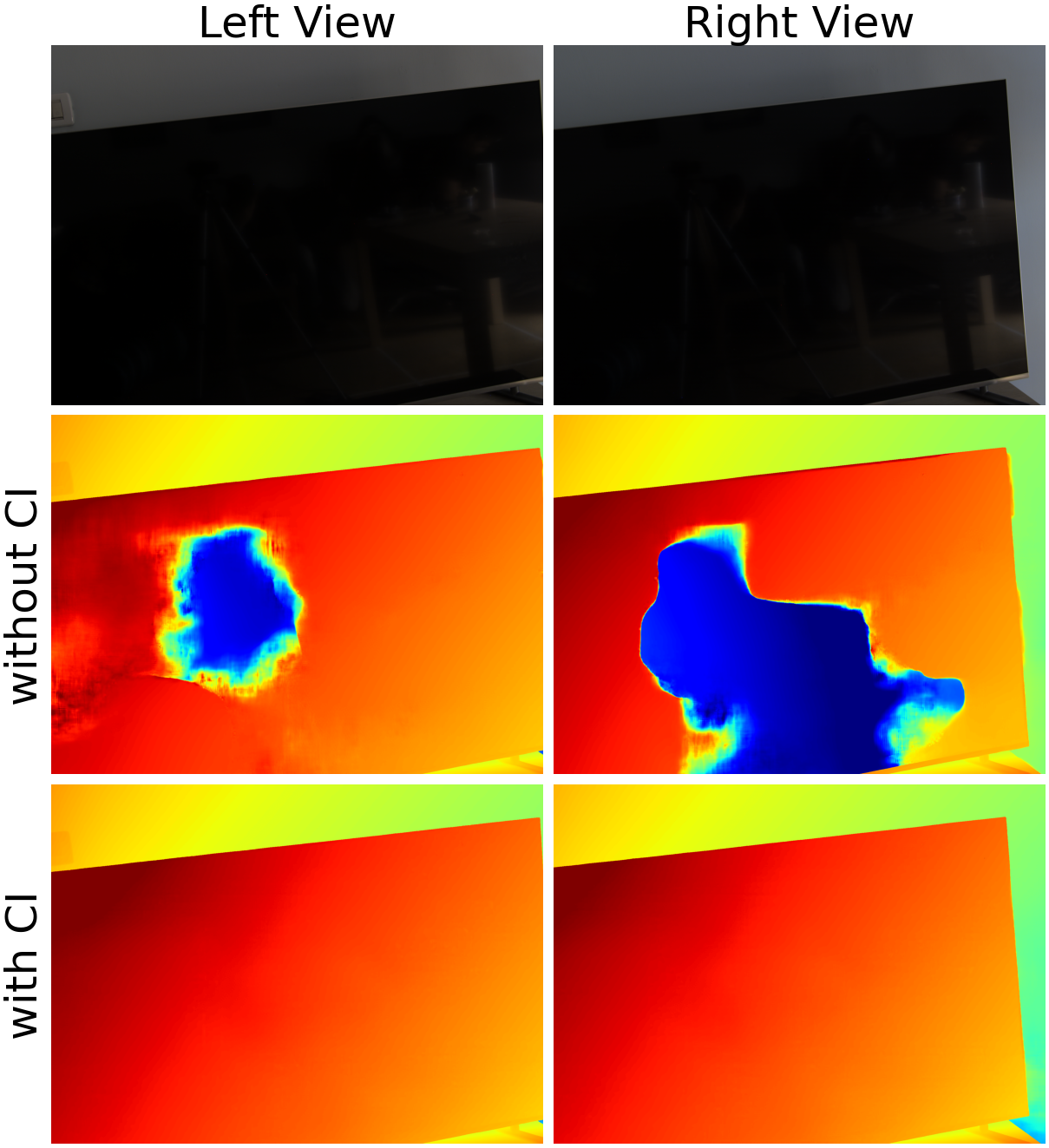}
    \caption{\textbf{Sample predictions with and without our proposed cross-injection (CI) extension.} CI can be added to any iterative stereo matcher (here FoundationStereo~\cite{wen2025foundationstereo}) and substantially improves left-right consistency between two otherwise independent inference passes (left example); it also reduces per-view failure modes (right example).
    Here, we show two examples from the Booster dataset~\cite{booster} where inconsistent predictions arise mainly on reflective and transparent surfaces. Additional visualizations are shown in Figs.~\ref{fig:eth3d-samples}, \ref{fig:booster-samples} and \ref{fig:kitti-samples}.
    }
    \label{fig:teaser}
\end{figure}

State-of-the-art stereo matchers, however, typically predict only the left (reference) disparity, a convention encouraged by benchmarks that score only the left view~\cite{middlebury,kitti12,kitti15,eth3d}.
In practice, the standard way to obtain the right disparity from such a network is to horizontally flip and swap the input pair, re-posing the right view as a left-view problem, and run the same weights a second time. The two passes are coupled only implicitly through the image content; nothing enforces that their outputs agree. As a result, they often diverge in textureless regions, at occlusion boundaries, and on transparent or reflective surfaces, where each pass locks onto a different plausible depth (Fig.~\ref{fig:teaser}, left example, middle row).

This is the gap we address in this paper: making the left and right predictions \emph{agree} without sacrificing single-view accuracy. Rather than a fundamental redesign (a symmetric, both-view architecture trained from scratch), we propose a lightweight extension that reuses the strong pretrained left-only matchers the field already provides. It attaches to any such matcher with a RAFT-style~\cite{teed2020raft,lipson2021raftstereo} refinement module and adds left-right consistency through fine-tuning alone, so it transfers to new base models as they appear.
In summary, our main contributions are as follows:
\begin{itemize}
    \item We propose \emph{cross-injection}, a lightweight extension for refinement-based left-only stereo matching models.
    Unlike prior in-network consistency mechanisms, we inject explicit disparity inconsistency directly into the GRU update. Our extension is zero-initialized, leaving the pretrained matcher's predictions unchanged at the start of fine-tuning, so the coupling is learned entirely by fine-tuning.
    \item We supervise this coupling with a left-right consistency loss under per-pixel bidirectional gradient routing, and show it preserves sharp edges better than a naive \emph{symmetric} one.
    \item We demonstrate that our method substantially improves left-right consistency for two state-of-the-art stereo matching architectures~\cite{wen2025foundationstereo,min2025s2m2}.
    \item We show that CI is also at least as accurate as a matched fine-tuning control in $14$ of $15$ benchmark--model combinations, and clearly better in most, making it a worthwhile extension even if only left-view prediction is required.
\end{itemize}


\section{Related Work}
\label{sec:related}

\paragraph{Iterative stereo refinement.}
Most state-of-the-art stereo matchers refine disparity with a recurrent update operator, a paradigm descending from end-to-end cost-volume regressors~\cite{kendall2017gcnet,chang2018psmnet}. RAFT~\cite{teed2020raft} introduced the GRU update over a correlation volume, and RAFT-Stereo~\cite{lipson2021raftstereo} adapted it to rectified stereo; later work strengthens the matching volume together with the recurrent update operator~\cite{li2022crestereo,xu2023igev,chen2024mocha} or the update operator alone~\cite{wang2024selective}, while others scale the approach with monocular priors~\cite{jiang2025defom,wen2025foundationstereo,bartolomei2025stereoanywhere} and large synthetic data~\cite{wen2025foundationstereo,min2025s2m2}. All these stereo models predict only the left disparity; our method extends this class.

\paragraph{Both-view prediction and left-right consistency.}
Predicting both disparities with symmetric occlusion reasoning has a long history~\cite{marr1976cooperative,belhumeur1996bayesian,sun2005symmetric} and recurs in deep networks via symmetric occlusion CNNs~\cite{li2018symmnet}, bidirectional cross-attention transformers~\cite{li2021sttr,xu2023unimatch}, and parallax attention~\cite{wang2022pasmnet}. The left-right disparity consistency loss originates with Godard~\etal~\cite{godard2017monodepth}, building on photometric self-supervision~\cite{garg2016unsupervised}, and recurs in later self-supervised depth and stereo models~\cite{poggi2018trinocular,zhong2017sssmnet}; related methods instead use a left-right consistency check to select reliable proxy labels~\cite{zhou2017unsupervisedstereo,tosi2019monoresmatch,wang2021pvstereo}, and unsupervised adaptation of a pretrained matcher relies on confidence-filtered proxy labels or photometric losses~\cite{tonioni2017adaptation,tonioni2019madnet}. In unsupervised optical flow, the two flow directions likewise provide a forward-backward consistency loss~\cite{meister2018unflow} and occlusion masks~\cite{meister2018unflow,wang2018occaware}. These consistency objectives are usually self-supervised and symmetric.

\paragraph{Consistency inside the network.}
Classically, the left-right check is a post-hoc occlusion filter~\cite{zbontar2016mccnn,hirschmuller2008sgm,scharstein2002taxonomy,egnal2002halfocclusion}; learned variants instead use cross-view signals during estimation via feature constancy~\cite{liang2018featureconstancy}, warping-error refinement~\cite{chabra2019stereodrnet}, attention cost volumes~\cite{xu2022acvnet}, cross-view-completion pretraining~\cite{weinzaepfel2022croco,weinzaepfel2023crocov2}, or auxiliary cross-view geometric constraints~\cite{gong2024crossview}. Closest to our refiner are LRCR~\cite{jie2018lrcr} and the refine-and-distill model of Pilzer~\etal~\cite{pilzer2019refinedistill}, which also feed left-right disagreement back into refinement. LRCR, however, turns this comparison into a \emph{learned} soft-attention map that flags only \emph{where} to refine; Pilzer~\etal instead derive their inconsistency as an image-space cycle-reconstruction error in a monocular model and turn it into a self-distillation signal (a stop-gradient pulls a student sub-network toward the refined disparity). By contrast, we inject the \emph{explicit, signed} disparity inconsistency (where, by how much, and in which direction the views disagree) as a parameter-free cue into a GRU update through a zero-initialized encoder.

\paragraph{Stop-gradient consistency targets.}
A symmetric consistency loss makes each of two predictions a target for the other, so it can be minimized by co-adapting both toward a degenerate mutual agreement rather than by correcting either. Mutual learning~\cite{zhang2018dml} keeps this in check by also anchoring each peer network to ground-truth labels; a pure consistency objective has no such anchor, so the standard remedy is a stop-gradient. It prevents the analogous collapse in Siamese self-supervised learning~\cite{chen2021simsiam,grill2020byol} and underpins consistency-based semi-supervised learning via a detached teacher~\cite{tarvainen2017meanteacher}. The same one-directional, detached signal recurs in stereo and flow: self-training in optical flow, where confident predictions supervise unreliable ones~\cite{liu2019ddflow,liu2019selflow}, stereo-to-monocular distillation from a fixed, pretrained stereo teacher~\cite{guo2018distillstereo}, and the self-distillation of Pilzer~\etal~\cite{pilzer2019refinedistill}. We adopt the same principle, but apply it per pixel and in both directions within a single weight-shared model: each view's left-right consistency gradient updates only the other view, and the warp field is detached so the loss cannot be reduced by sliding the sampling location instead of correcting the disparity (Sec.~\ref{sec:ts-lr-consistency}).

\begin{figure*}
    \centering
\includegraphics[width=\linewidth]{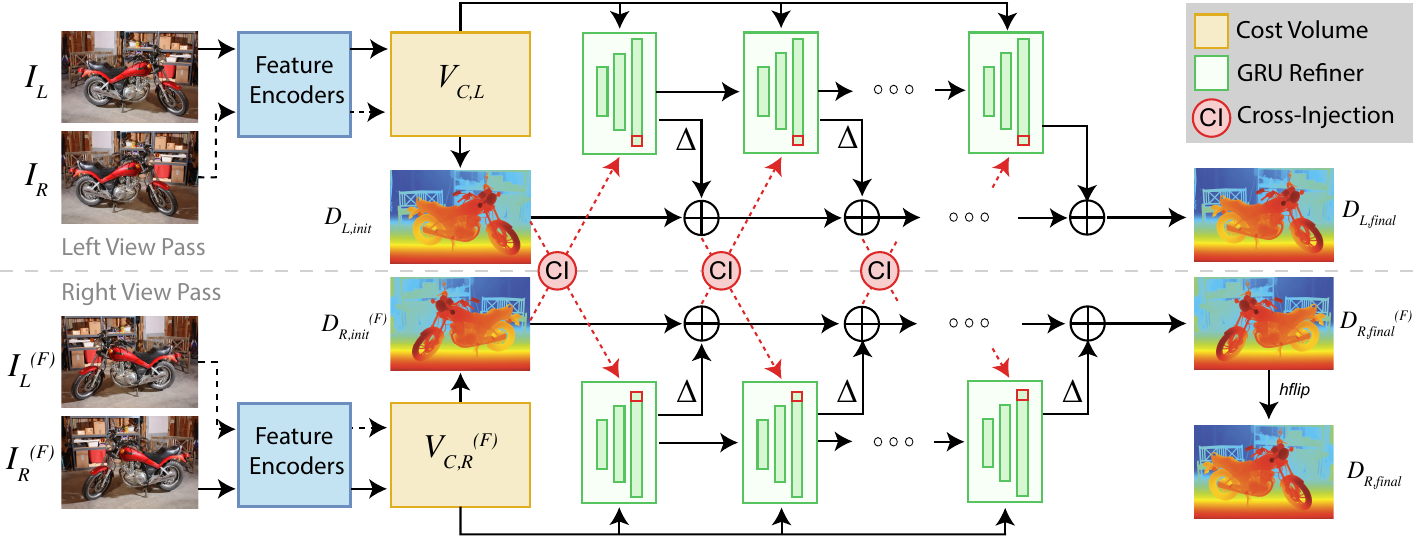}
    \caption{\textbf{Overview of the proposed cross-injection architecture.} Two refinement passes (the left pass and the flipped-swapped right pass), whose feature encoders and GRU share weights, each iteratively update their own disparity. At every step, each pass receives its own masked left-right inconsistency with respect to the other view (Fig.~\ref{fig:ci}, Eq.~\ref{eq:ci}) as an additional input cue, injected through a zero-initialized encoder. Because the encoder emits zero at initialization, augmenting an already-trained refiner with cross-injection reproduces its original (uncoupled) behavior exactly, and the cross-view correction is learned by fine-tuning.}
    \label{fig:arch}
\end{figure*}

\begin{figure}
    \centering
\includegraphics[width=0.98\linewidth]{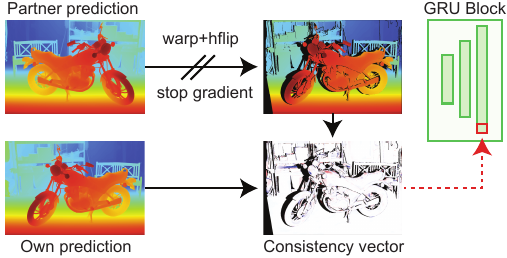}
    \caption{\textbf{Illustration of the proposed cross-injection mechanism.} The detached partner prediction is horizontally flipped and warped into the frame of the own prediction, using the own disparity as the shift field (Eq.~\ref{eq:ci}), and subtracted from the own prediction. The resulting signed inconsistency (``consistency vector'') is passed explicitly to the refiner's GRU block. We do this with a tensor of shape $(B,2,H,W)$:
    the two channels are (i) the signed inconsistency with zeros at non-co-visible pixels, and (ii) the binary co-visibility mask derived from warping. This way, the model can distinguish non-visibility from perfect consistency.}
    \label{fig:ci}
\end{figure}

\section{Approach}
\label{sec:methods}

Our main contribution is a lightweight \emph{cross-injection} refiner (Sec.~\ref{sec:cross-injection}) that couples the left and right disparity predictions during iterative refinement; we train it with a \emph{left-right consistency loss} (Sec.~\ref{sec:ts-lr-consistency}) that supervises the same coupling at the output, using bidirectional gradient routing. The two are complementary: the refiner consumes the per-pixel left-right inconsistency as an input cue, while the loss penalizes its magnitude during training, so the same geometric quantity is addressed by
both the architecture and the objective.

\subsection{Stereo Refinement via Cross-Injection}
\label{sec:cross-injection}

Our extension applies to any iterative (refinement-based) stereo matcher trained to predict left disparities only~\cite{lipson2021raftstereo,min2025s2m2,wen2025foundationstereo}. The right disparity is obtained as in Sec.~\ref{sec:intro} by flipping and swapping the input pair and flipping the prediction back. Each pass refines its disparity with a RAFT-style~\cite{teed2020raft,lipson2021raftstereo} convolutional GRU that gathers correlation features at the current estimate and emits an increment; the passes share weights but not state and interact only through the correlation volume, so agreement would be encouraged only indirectly by the disparity loss (Eq.~\ref{eq:disp}). Cross-injection instead tells each pass, at every step and pixel, where and by how much it disagrees with the other view, so it can act on the disagreement during refinement.

Writing $\mathbf{p}=(x,y)$ for a pixel of the $H\!\times\!W$ grid, each network pass starts from an initial disparity $D^{(0)}\in\mathbb{R}^{H\times W}$  and refines it over $K$ steps into $D^{(k)}$, $k=1,\dots,K$, giving $D_L^{(k)}$ and $D_R^{(k)}$ for the two passes (iteration index dropped where unambiguous). Disparities are non-negative; a left pixel $(x,y)$ corresponds to the right pixel $(x-D_L,\,y)$ and a
right pixel $(x,y)$ to the left pixel $(x+D_R,\,y)$.

\paragraph{Consistency estimation.}
We compare the two views in disparity space with a horizontal bilinear sampler that warps a source map $S$ along a shift field $\delta$ with sign $s\in\{+1,-1\}$,
\begin{equation}
  \big[\mathcal{S}_{s}(S,\delta)\big](x,y) \;=\;
  S\big(x + s\,\delta(x,y),\; y\big),
  \label{eq:sampler}
\end{equation}
with out-of-grid samples set to zero. Sampling the other view's disparity in the current view's frame and subtracting gives the per-pixel left-right \emph{inconsistency}
\begin{equation}
\begin{aligned}
  c_L \;=\; D_L - \mathcal{S}_{-}(D_R,\,D_L),
  \\[2pt]
  c_R \;=\; D_R - \mathcal{S}_{+}(D_L,\,D_R),
  \label{eq:ci}
\end{aligned}
\end{equation}
which is zero where the views agree. It is only meaningful at co-visible pixels, so we restrict it with a mask of non-occluded pixels that have an in-bound match,
\begin{equation}
  m_L(\mathbf{p}) =
    \ind\!\big[\,0 \le x - D_L(\mathbf{p}) < W\,\big]\; V_L(\mathbf{p}),
  \label{eq:covis}
\end{equation}
and symmetrically for $m_R$. Here $V_L\in\{0,1\}$ flags non-occluded pixels, obtained from $D_L$ by a standard single-view uniqueness (z-buffer) test~\cite{egnal2002halfocclusion,sun2005symmetric}: each pixel is forward-warped to the other view and, on collision, only the nearest (largest-disparity) pixel survives; a 1-pixel erosion of this mask discards unreliable pixels near object boundaries. During cross-injection (Fig.~\ref{fig:ci}), the mask is recomputed at each step from the current detached disparity and treated as constant.

\paragraph{Cross-injection.}
The refiner consumes the masked inconsistency as an extra input to the GRU's finest-resolution update features. The coupling is bidirectional and symmetric: each view receives its own inconsistency at every step. 
The partner view's disparity enters this cue with a stop-gradient, so cross-injection carries no gradient between the two passes while each pass still receives the other's disparity as an input cue (Fig.~\ref{fig:ci}). The mask $m$ itself is appended as a second channel so the model can distinguish ``not visible in both views'' from ``views agree,'' which would otherwise both appear as zeros. The masked inconsistency tensor of shape $(B,2,H,W)$ is concatenated to the last layer of the GRU refinement block through a zero-initialized encoder. In the case of FoundationStereo, this increases the layer's feature dimension from $384$ to $386$. Appendix~\ref{sec:lrcr} ablates the three choices that define this cue (signed vs.\ unsigned, explicit vs.\ learned, and masked vs.\ unmasked) and finds that the sign is what matters: only the signed residual substantially improves left-right consistency, while the mask changes little.

Three properties make cross-injection attractive. It is \emph{lightweight}: the inconsistency is computed without learned parameters (a warp, a subtraction, and a mask) and injected through a small zero-initialized encoder (only $0.01\%$ of the FoundationStereo base model's parameters). It \emph{couples} two otherwise independent passes: the left and flipped-swapped right runs interact only through this signal, turning two single-view refinements into a binocular one. And it can be \emph{added after training}: the zero-initialized encoder emits zero at initialization, so attaching cross-injection leaves the pretrained model's predictions unchanged, and the coupling is learned by fine-tuning.

\subsection{Left-Right Consistency Loss}
\label{sec:ts-lr-consistency}

Cross-injection lets the refiner read the left-right inconsistency $c$ (Eq.~\ref{eq:ci}); the loss closes the loop by penalizing its magnitude during training~\cite{godard2017monodepth}. A physically correct stereo solution must satisfy the geometric constraint that two pixels in correspondence carry the same disparity ($c{=}0$); turning this into a training signal (i)~couples the two model passes, (ii)~lets a view with ground truth supervise a view that has none (e.g.\ left-only annotations), and (iii)~regularizes the prediction beyond what the supervised disparity term provides. The plain constraint, however, is \emph{symmetric}: it pulls both predictions toward each other, so it can be satisfied by co-adapting both toward a mutually consistent but inaccurate compromise rather than by correcting either. We restrict the loss to co-visible pixels with the mask $m$ (Eq.~\ref{eq:covis}) and route each direction's gradient through a per-pixel stop-gradient.

\paragraph{Base consistency term.}
Let $\mu(w,r)$ denote the mask-weighted mean of a residual $r$ under a
non-negative per-pixel weight $w$,
\begin{equation}
  \mu(w,r) \;=\;
  \frac{\sum_{\mathbf{p}} w(\mathbf{p})\,r(\mathbf{p})}
       {\sum_{\mathbf{p}} w(\mathbf{p})},
  \label{eq:masked-mean}
\end{equation}
for a non-empty mask $w$. The naive, symmetric consistency loss is the masked mean inconsistency magnitude, averaged over the two views,
\begin{equation}
  \ell^{(k)}_{\mathrm{base}} \;=\;
  \tfrac{1}{2}\Big[\,
    \mu\big(m_L,\; |c_L|\big) +
    \mu\big(m_R,\; |c_R|\big)
  \Big].
  \label{eq:base}
\end{equation}
Here the gradient of $c$ (Eq.~\ref{eq:ci}) flows into both views: each residual updates the unwarped map, the sampled values, and the sampling location of the warp, pulling the two endpoints of every correspondence toward each other. This is the variant we ablate against: it improves left-right agreement, but agreement can be met by co-smoothing both predictions, which blurs depth boundaries and occluded regions (Sec.~\ref{sec:results}).

\paragraph{Bidirectional gradient routing.}
In each direction we detach one view together with the warp field, so that the only gradient path is through the bilinearly sampled
values of the other view,
\begin{equation}
\begin{aligned}
  \tilde{c}_L &= \sg[D_L] - \mathcal{S}_{-}\!\big(D_R,\, \sg[D_L]\big),
  \label{eq:residual-ts}\\
  \tilde{c}_R &= \sg[D_R] - \mathcal{S}_{+}\!\big(D_L,\, \sg[D_R]\big),
\end{aligned}
\end{equation}
where $\sg[\cdot]$ is the stop-gradient. In $\tilde{c}_L$ the left map is held fixed and the gradient updates only the right map; $\tilde{c}_R$ is its mirror image. The loss combines both directions, each restricted to its co-visible pixels,
\begin{equation}
  \ell^{(k)}_{\mathrm{lrc}} \;=\;
  \tfrac{1}{2}\Big[\,
    \underbrace{\mu\big(m_L,\; |\tilde{c}_L|\big)}
      _{\text{left teaches right}}
    \;+\;
    \underbrace{\mu\big(m_R,\; |\tilde{c}_R|\big)}
      _{\text{right teaches left}}
  \,\Big].
  \label{eq:ts}
\end{equation}

The reason to detach one side is the degeneracy noted in Sec.~\ref{sec:related}: a symmetric agreement loss makes each prediction a target for the other, so it can be lowered by co-adapting both toward a mutually consistent compromise rather than by correcting either. Where the data does not pin down the predictions, two maps can agree simply by being wrong in the same way. Detaching one side turns this mutual objective into a one-directional regression: the loss can then be lowered only by moving the updated view onto the detached view's current value, not by degrading both. Because the routing is applied in both directions, neither view is privileged: a symmetric, mutual arrangement in the spirit of deep mutual learning~\cite{zhang2018dml}, but driven by a per-pixel stop-gradient rather than peer mimicry. The stop-gradient also covers $D_L$ in its second role as the warp index $x-D_L$ that sets where the partner map is read, since a differentiable index would otherwise let the loss lower the residual by sliding the read location onto a coincidentally matching pixel rather than by correcting depth.

The supervised disparity term (Eq.~\ref{eq:disp}) keeps the predictions from collapsing globally and, on samples where only one view is labeled, anchors that view, so that the consistency term moves mainly the unlabeled one into agreement; but where that supervision is weak (depth boundaries, occlusions, and the unlabeled view) the symmetric loss still drifts toward the blurred consensus, which we confirm in Sec.~\ref{sec:results}.

\paragraph{Full objective.}
The supervised disparity loss is the standard masked $L_1$ error against ground truth, pooled over both views wherever ground truth exists,
\begin{equation}
  \ell^{(k)}_{\mathrm{disp}} = \mu\big(m^{\mathrm{gt}},\,|D^{(k)}-\hat D|\big),
  \label{eq:disp}
\end{equation}
where $m^{\mathrm{gt}}$ is a binary mask selecting pixels with valid ground truth. Some training datasets label both views and others only one; on the latter, the unlabeled view is supervised by the consistency term alone.
For FoundationStereo, both terms are applied at every refinement step:
\begin{equation}
  \mathcal{L} =
    \sum_{k=1}^{K}\gamma^{\,K-k}\, \left( \ell^{(k)}_{\mathrm{disp}}
    \;+\; \lambda\,\ell^{(k)}_{\mathrm{lrc}} \right),
  \label{eq:full}
\end{equation}
where $\lambda=0.5$ weights the consistency term (swept in Appendix~\ref{sec:lambda}) and the per-step decay $\gamma\in(0,1)$ follows the base models' original implementations~\cite{wen2025foundationstereo,min2025s2m2}. For \SSMM, whose intermediate predictions are at quarter resolution, the disparity term is the model's native multi-resolution sequence loss and the consistency term is applied to the final full-resolution prediction only.

\section{Experiments}
\label{sec:results}

\begin{table*}[t]
\centering
\caption{\textbf{Ablation experiment for component validation.} All runs are fine-tuning runs based on FoundationStereo. Checkmarks indicate which components each run uses: LR-consistency loss (LR), cross-injection (CI) refiner, and bidirectional stop-gradient routing (SG). All metrics are reported as mean$\pm$std of six checkpoints created during the last 10k steps of training.
\textbf{Findings}: LR consistency improves with the loss alone (b), only slightly with CI alone (c), and substantially when both are combined (d). However, CI with a symmetric consistency loss leads to (slight) oversmoothing on edges and degradation of occluded predictions. SG largely recovers this regression at the cost of slightly lower LR consistency.
\textbf{Annotation}: Per column, \textbf{bold} = best; \colorbox{lred}{red}/\colorbox{lgreen}{green} mark the most salient degradation/improvement trends in the metric category compared to the baseline.}
\label{tab:ablation}
\begin{adjustbox}{max width=\textwidth}
\begin{tabular}{@{}c@{\;}c@{\;}c@{\;}c|ccc|cccc|ccc}
\toprule
\multicolumn{4}{@{}c|}{Ablation} & \multicolumn{3}{c|}{Disparity Error} & \multicolumn{4}{c|}{Left-Right Consistency} & \multicolumn{3}{c}{Edge/Occlusion Error} \\
\midrule
& LR & CI & SG & \shortstack{Middleb.-H\\BP-2} & \shortstack{ETH3D\\BP-1} & \shortstack{KITTI-15\\D1} & \shortstack{Middleb.-H\\LR-BP-1} & \shortstack{ETH3D\\LR-BP-1} & \shortstack{KITTI-15\\LR-BP-1} & \shortstack{Booster-H\\LR-BP-1} & \shortstack{Middleb.-H\\Edg. BP-0.5} & \shortstack{Middleb.-H\\Occ. BP-0.5} & \shortstack{Booster-H\\Occ. BP-0.5} \\
\midrule
(a) &  &  &  & 2.78$\pm$0.09 & 0.72$\pm$0.07 & 2.81$\pm$0.02 & 9.98$\pm$0.19 & 5.12$\pm$0.10 & 10.31$\pm$0.19 & 7.76$\pm$0.15 & \textbf{32.25$\pm$0.03} & \textbf{33.39$\pm$0.51} & \textbf{51.12$\pm$0.50} \\ \midrule
(b) & \checkmark &  &  & \textbf{2.66$\pm$0.04} & 0.69$\pm$0.05 & 2.75$\pm$0.02 & 9.20$\pm$0.07 & 4.63$\pm$0.04 & 5.63$\pm$0.04 & 6.26$\pm$0.17 & 33.11$\pm$0.09 & 33.68$\pm$0.30 & 51.28$\pm$0.39 \\
(c) &  & \checkmark &  & 2.72$\pm$0.04 & 0.68$\pm$0.02 & 2.81$\pm$0.02 & 9.63$\pm$0.09 & 4.81$\pm$0.30 & 9.67$\pm$0.19 & 6.78$\pm$0.11 & 32.34$\pm$0.05 & 33.66$\pm$0.65 & 51.26$\pm$0.43 \\
(d) & \checkmark & \checkmark &  & 2.69$\pm$0.03 & 0.65$\pm$0.01 & 2.74$\pm$0.03 & \cellcolor{lgreen}{\textbf{8.83$\pm$0.02}} & \cellcolor{lgreen}{\textbf{2.92$\pm$0.03}} & \cellcolor{lgreen}{\textbf{4.33$\pm$0.06}} & \cellcolor{lgreen}{\textbf{3.23$\pm$0.06}} & \cellcolor{lred}{33.30$\pm$0.17} & \cellcolor{lred}{34.60$\pm$0.25} & \cellcolor{lred}{52.35$\pm$0.28} \\
(e) & \checkmark & \checkmark & \checkmark & 2.68$\pm$0.03 & \textbf{0.58$\pm$0.01} & \textbf{2.71$\pm$0.02} & \cellcolor{lgreen}{9.06$\pm$0.02} & \cellcolor{lgreen}{3.24$\pm$0.02} & \cellcolor{lgreen}{5.44$\pm$0.07} & \cellcolor{lgreen}{3.64$\pm$0.07} & 32.29$\pm$0.05 & 33.58$\pm$0.32 & 51.37$\pm$0.39 \\
\bottomrule
\end{tabular}
\end{adjustbox}
\end{table*}

\begin{table*}[t]
\centering
\caption{\textbf{Cross-injection improvements generalize to different base models.} For each base model we report three rows: the public checkpoint, a \emph{matched fine-tune} under the identical recipe but \emph{without} cross-injection (``+\,FT (no CI)''), and our cross-injection fine-tune. The middle row is the control: it isolates what CI contributes from what simply training longer on our data mixture contributes. \textbf{Findings:} against the matched control, CI improves left-right consistency on \emph{every} benchmark for all three base models, with the largest relative effect on Booster, which is built around transparent and specular surfaces (for \SSMM-XL, KITTI-12 is marginally larger). Disparity error (both views where GT is available) also improves against the control on all but one of the fifteen accuracy cells (\SSMM-L Booster-H). Bold: best per base model.}
\label{tab:main_metrics}
\begin{adjustbox}{max width=\textwidth}
\begin{threeparttable}
\begin{tabular}{lcccccccccc}
\toprule
 & \multicolumn{2}{c}{Middlebury-H} & \multicolumn{2}{c}{ETH3D} & \multicolumn{2}{c}{KITTI-12} & \multicolumn{2}{c}{KITTI-15} & \multicolumn{2}{c}{Booster-H} \\
\cmidrule(lr){2-3} \cmidrule(lr){4-5} \cmidrule(lr){6-7} \cmidrule(lr){8-9} \cmidrule(lr){10-11}
Model & BP-2 & LR-BP-1 & BP-1 & LR-BP-1 & D1 & LR-BP-1 & D1 & LR-BP-1 & BP-2 & LR-BP-1 \\
\midrule
FoundationStereo & 2.77 & 9.89 & 0.65 & 5.59 & 2.73 & 12.20 & 2.88 & 10.23 & 10.70 & 8.84 \\
FoundationStereo + FT (no CI) $^{\dagger}$ & 2.78 & 10.00 & 0.68 & 4.99 & 2.64 & 11.78 & 2.80 & 10.12 & 9.27 & 7.86 \\
FoundationStereo + CI $^{\dagger}$ & \textbf{2.68} & \textbf{9.03} & \textbf{0.61} & \textbf{3.26} & \textbf{2.49} & \textbf{7.09} & \textbf{2.69} & \textbf{5.34} & \textbf{9.24} & \textbf{3.60} \\
\midrule
\SSMM-L $^{\ddag}$ & 4.11 & 10.05 & 0.92 & 4.24 & 3.18 & 12.44 & 3.12 & 7.66 & 14.68 & 16.27 \\
\SSMM-L + FT (no CI) & 3.63 & 10.08 & 0.89 & 3.77 & 2.35 & 9.81 & 2.57 & 6.85 & \textbf{11.17} & 11.60 \\
\SSMM-L + CI & \textbf{3.47} & \textbf{9.51} & \textbf{0.64} & \textbf{3.03} & \textbf{2.16} & \textbf{7.27} & \textbf{2.44} & \textbf{5.43} & 11.86 & \textbf{7.90} \\
\midrule
\SSMM-XL $^{\ddag}$ & 2.98 & 9.58 & 0.64 & 5.05 & 4.12 & 16.26 & 2.97 & 8.56 & 7.00 & 8.31 \\
\SSMM-XL + FT (no CI) & 2.95 & 9.42 & 0.63 & 4.02 & 2.10 & 11.58 & 2.31 & 7.37 & 6.54 & 5.86 \\
\SSMM-XL + CI & \textbf{2.81} & \textbf{9.05} & \textbf{0.53} & \textbf{3.21} & \textbf{2.00} & \textbf{7.58} & \textbf{2.24} & \textbf{5.38} & \textbf{6.53} & \textbf{3.92} \\
\bottomrule
\end{tabular}
\begin{tablenotes}
\item {\footnotesize $\dagger$: As Tab.~\ref{tab:ablation} (a) and (e), but the final checkpoint instead of the six-checkpoint mean, for comparability with external models.
}
\item {\footnotesize $\ddag$: The \SSMM checkpoints are those published by the authors on \href{https://github.com/junhong-3dv/s2m2}{GitHub}. The models used to report the SOTA benchmark metrics on Middlebury are not released.}
\end{tablenotes}
\end{threeparttable}
\end{adjustbox}
\end{table*}

We validate our method in two parts: a component ablation on FoundationStereo (Sec.~\ref{sec:ablation}) that isolates the effect of each component, and a generalization study (Sec.~\ref{sec:results-generalization}) that applies the full method to additional pretrained stereo matching models. In both we report single-view disparity accuracy alongside left-right consistency, evaluated zero-shot on benchmarks unseen during training.

\paragraph{Metrics.}
We report bad-pixel rates BP-$x$ (percentage of valid pixels whose disparity error exceeds $x$\,px) and, on KITTI, the standard D1 error. Binocular agreement is measured by LR-BP-1: the percentage of left-view pixels whose predicted match falls inside the right image and at which the left disparity and the right disparity sampled at that match disagree by more than $1$\,px, computed per image and averaged over images; this requires no ground truth. Occluded pixels are not excluded: for a correct prediction they disagree by roughly the occlusion width, so they add a floor close to the occluded fraction of the image (for FoundationStereo on Middlebury-H, $9.89$ overall against $2.00$ on ground-truth non-occluded pixels), which compresses relative differences on Middlebury. To localize where the consistency coupling helps or hurts, we additionally score two ground-truth regions of interest at the sub-pixel BP-0.5 threshold. \emph{Edges} are pixels within $2$\,px of a depth discontinuity, itself defined as a disparity jump larger than $2$\,px between adjacent pixels with valid ground truth, the region where oversmoothing and edge fattening concentrate. \emph{Occlusions} are valid pixels that fail a left-right ground-truth consistency check (forward-matched disparities disagreeing by more than $1$\,px), and are therefore reported only where both views have ground truth. Disparity and region-of-interest metrics are averaged over both views on datasets with two-sided ground truth (Middlebury, Booster) and over the reference view otherwise.

\subsection{Ablation Study}
\label{sec:ablation}
To assess the contribution of the components proposed in Sec.~\ref{sec:methods},
we ablate our method with a cumulative study, starting from the left-only pretrained FoundationStereo~\cite{wen2025foundationstereo} checkpoint and switching on one component at a time during fine-tuning (Tab.~\ref{tab:ablation}). Beyond disparity accuracy and left-right consistency, we report targeted region-of-interest metrics at depth boundaries (edges) and in occluded areas to surface any regressions introduced by the consistency coupling. All variants share an identical schedule and the same subset of FoundationStereo's original training data (Appendix~\ref{sec:training_details}, Supplementary Materials). We evaluate cross-dataset generalization on five unseen benchmarks: Middlebury~\cite{middlebury}, ETH3D~\cite{eth3d}, KITTI-12~\cite{kitti12}, KITTI-15~\cite{kitti15}, and Booster~\cite{booster}. To reduce per-step noise we report the mean and standard deviation over six checkpoints from the final $10$k steps rather than a single snapshot.

Starting from a baseline~(a) that predicts both views without any left-right coupling, we add the left-right consistency loss~(b), then the cross-injection refiner~(c), both jointly (d), and finally the bidirectional gradient routing~(e). The consistency loss and the refiner together drive the two views into much closer agreement (d) (see green highlights in Tab.~\ref{tab:ablation}), but this symmetric coupling oversmooths depth boundaries and occluded regions (red highlights).
The refiner alone~(c) improves consistency only modestly (Booster-H LR-BP-1 $7.8\!\rightarrow\!6.8$), presumably because nothing in the objective rewards acting on the cue; together with the loss~(d), the disagreement is more than halved ($3.2$), well beyond either component alone, consistent with their complementary roles (Sec.~\ref{sec:methods}).
The bidirectional gradient routing (e) removes this side effect: all three region-of-interest columns return to within $0.25$ points of the uncoupled baseline (inside the checkpoint-to-checkpoint spread in every case), against the consistent $1.0$ to $1.25$ point regression of the symmetric loss (d), while the left-right consistency gains are largely preserved; this constitutes our full method.
Notably, even row~(b), the loss alone with no architectural change at inference, already improves left-right consistency, most clearly on KITTI-15 ($10.3\!\rightarrow\!5.6$), so coupling the two views in one backward pass is itself a useful regularizer.

\subsection{Generalization to Other Base Models}
\label{sec:results-generalization}

Having validated each component on FoundationStereo, we now ask whether CI is a general recipe for iterative stereo matching or an artifact of a single architecture. We apply CI to \SSMM~\cite{min2025s2m2}, one of the few top-ranked methods on the Middlebury leaderboard that releases a public checkpoint and uses a GRU-style recurrent refiner. Beyond availability, \SSMM is a deliberately demanding test case because it relies far less on iterative refinement, running only $3$ update iterations instead of $32$, which leaves cross-injection a much smaller window in which to couple the two views. We evaluate the two publicly released sizes, \SSMM-L and \SSMM-XL, fine-tuning each from its released checkpoint with the same cross-injection recipe as in Sec.~\ref{sec:ablation}; per-model training configs are given in Appendix~\ref{sec:training_details}.
We fine-tune each model only on a subset of \SSMM's own original training data, introducing no data \SSMM was not already trained on, so that the gains reflect the method rather than additional supervision.

Tab.~\ref{tab:main_metrics} shows that cross-injection reproduces the FoundationStereo pattern on both \SSMM\ sizes. Against the matched control (the identical fine-tuning recipe with cross-injection and the consistency loss switched off), left-right consistency improves on every benchmark for both models, most strongly on Booster, whose transparent and specular surfaces are a classic source of left-right disagreement ($11.6\!\rightarrow\!7.9$ for \SSMM-L and $5.9\!\rightarrow\!3.9$ for \SSMM-XL, about a third), and on KITTI-12 ($9.8\!\rightarrow\!7.3$ and $11.6\!\rightarrow\!7.6$). Single-view accuracy does not suffer: disparity error is held or improved on all but one benchmark across both sizes, often substantially (ETH3D BP-1 $0.89\!\rightarrow\!0.64$ for \SSMM-L); the sole regression is a $0.7$-point rise on Booster BP-2 for \SSMM-L. That these improvements transfer to models running an order of magnitude fewer refinement steps, and under a strictly smaller training set than the original, suggests that cross-injection couples the two views through a mechanism that is not specific to one architecture or data regime.

\subsection{Qualitative Results}

\begin{figure*}[p]
      \centering
      \begin{subfigure}[b]{0.31\textwidth}
          \centering
          \includegraphics[width=\textwidth]{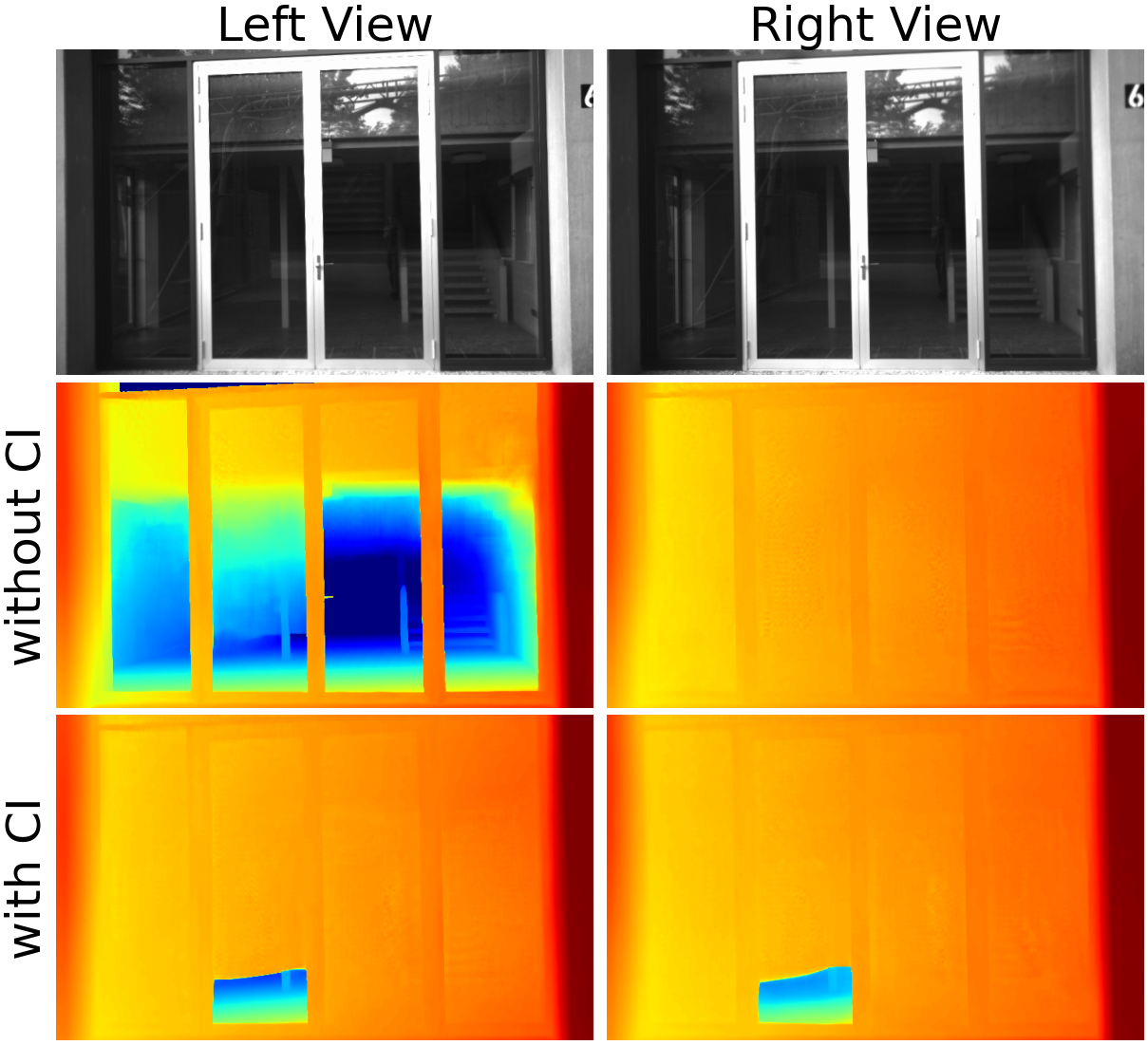}
          \caption{electro\_2s}
          \label{fig:eth3d-a}
      \end{subfigure}
      \hfill
      \begin{subfigure}[b]{0.35\textwidth}
          \centering
          \includegraphics[width=\textwidth]{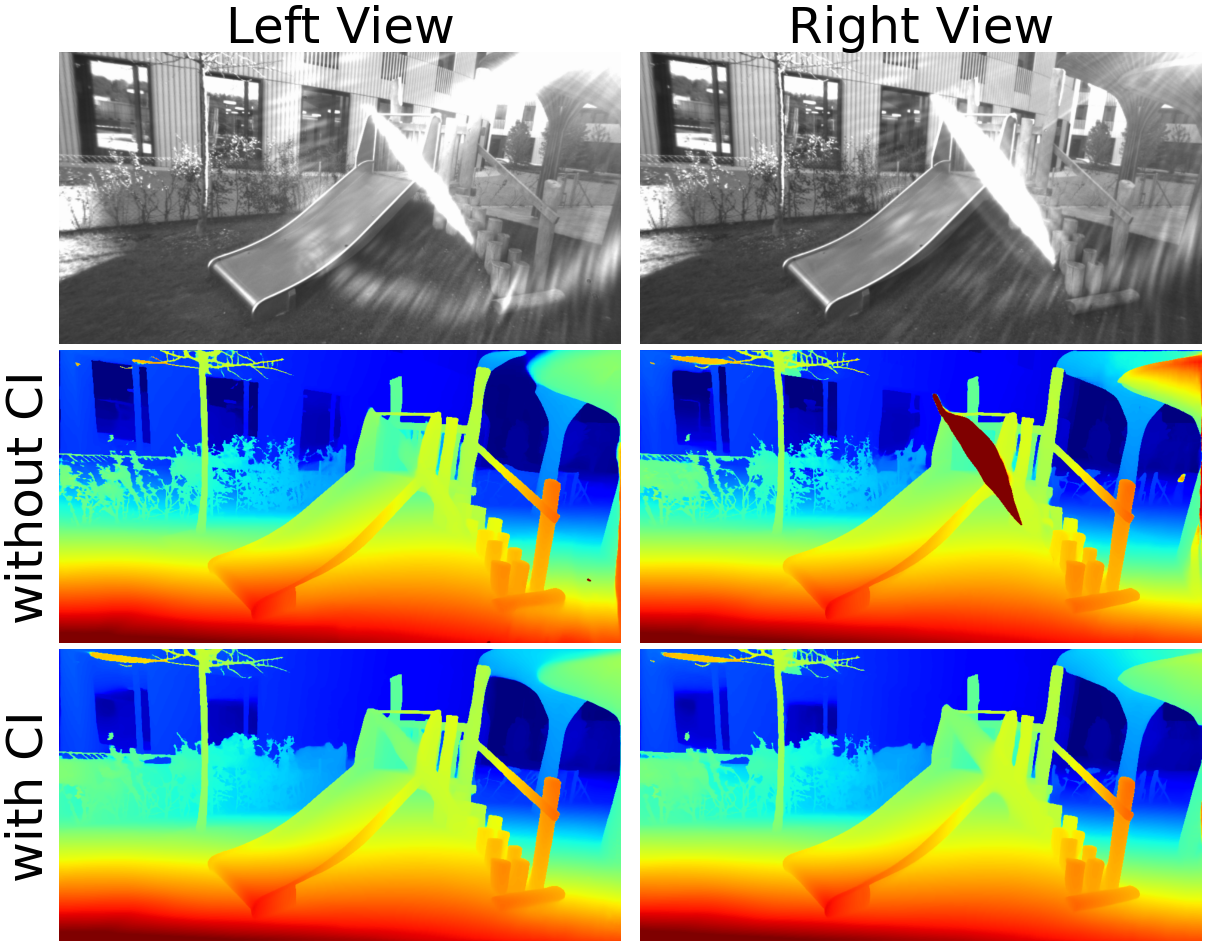}
          \caption{playground\_2l}
          \label{fig:eth3d-b}
      \end{subfigure}
      \hfill
      \begin{subfigure}[b]{0.31\textwidth}
          \centering
          \includegraphics[width=\textwidth]{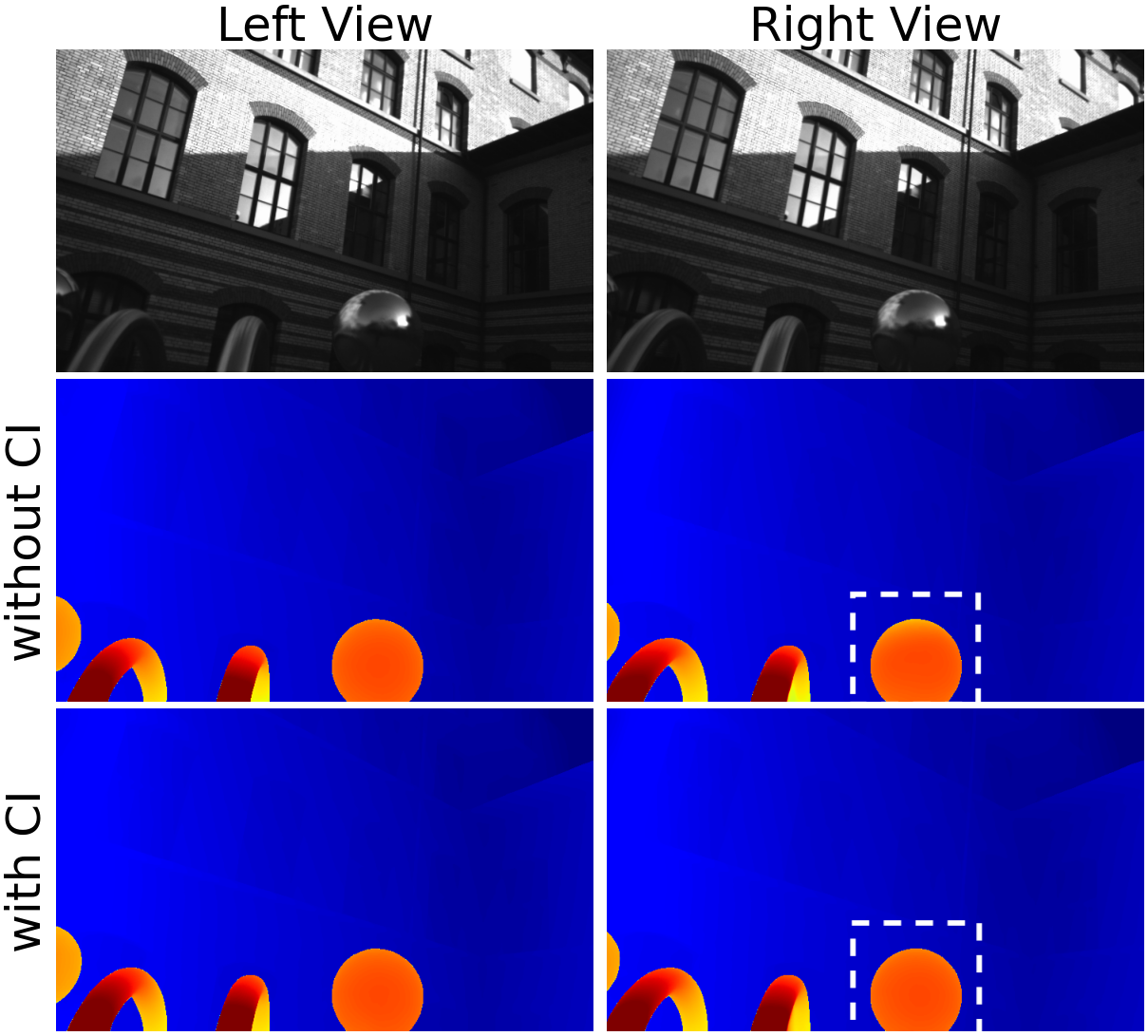}
          \caption{facade\_1s}
          \label{fig:eth3d-c}
      \end{subfigure}
      \caption{FoundationStereo zero-shot predictions on selected \textbf{ETH3D}~\cite{eth3d} samples with and without our cross-injection method. Even though we are not addressing the ambiguity for transparent objects directly, left/right predictions are more consistent in which visible layer to predict (a). The CI model also successfully corrects major (b) and minor (c) disagreements between the views. Background shrubs in (b) lose some degree of detail.}
      \label{fig:eth3d-samples}
  \end{figure*}

\begin{figure*}[p]
      \centering
      \begin{subfigure}[b]{0.245\textwidth}
          \centering
          \includegraphics[width=\textwidth]{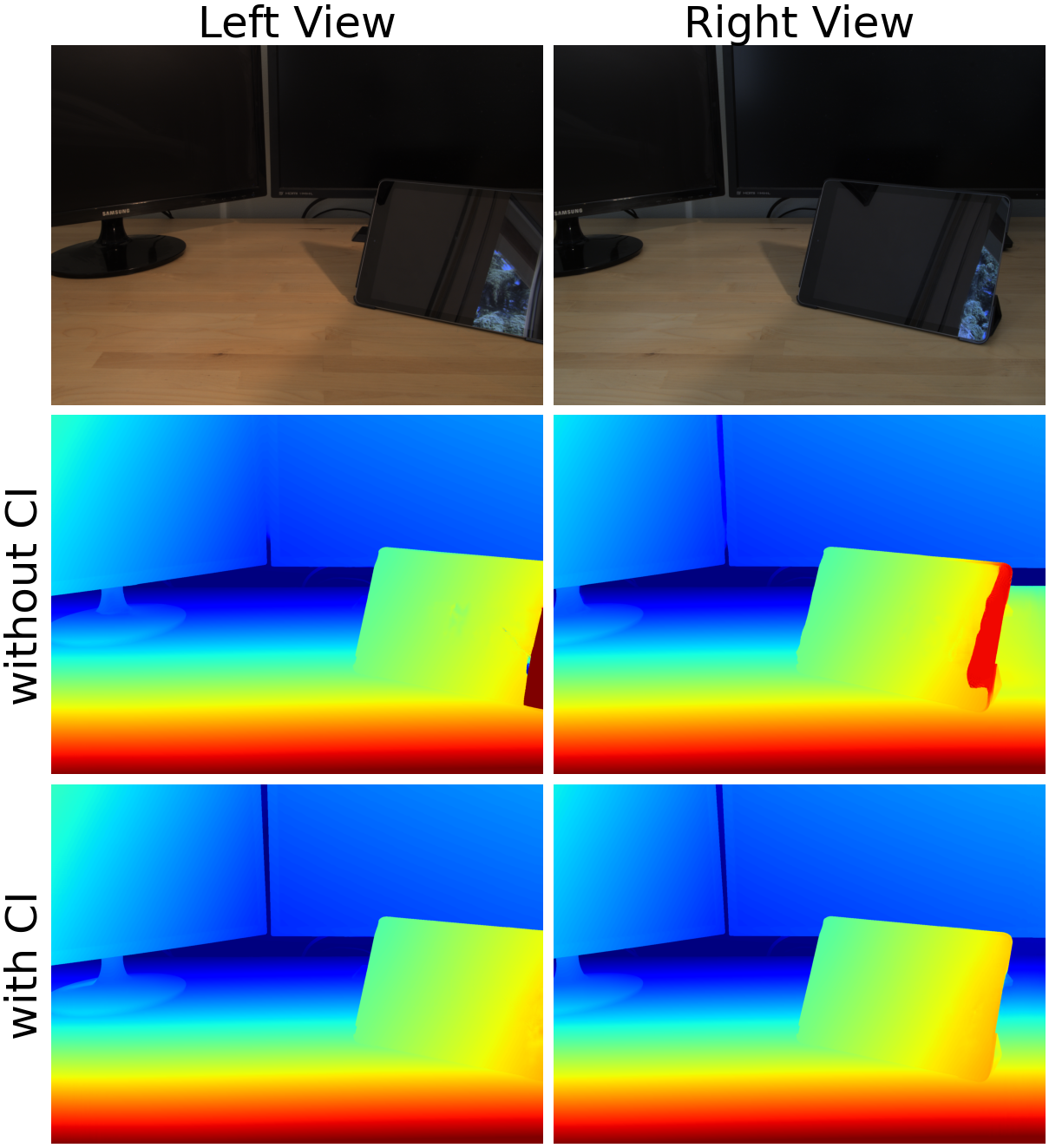}
          \caption{Tablet\_im3}
          \label{fig:booster-c}
      \end{subfigure}
      \hfill
      \begin{subfigure}[b]{0.245\textwidth}
          \centering
          \includegraphics[width=\textwidth]{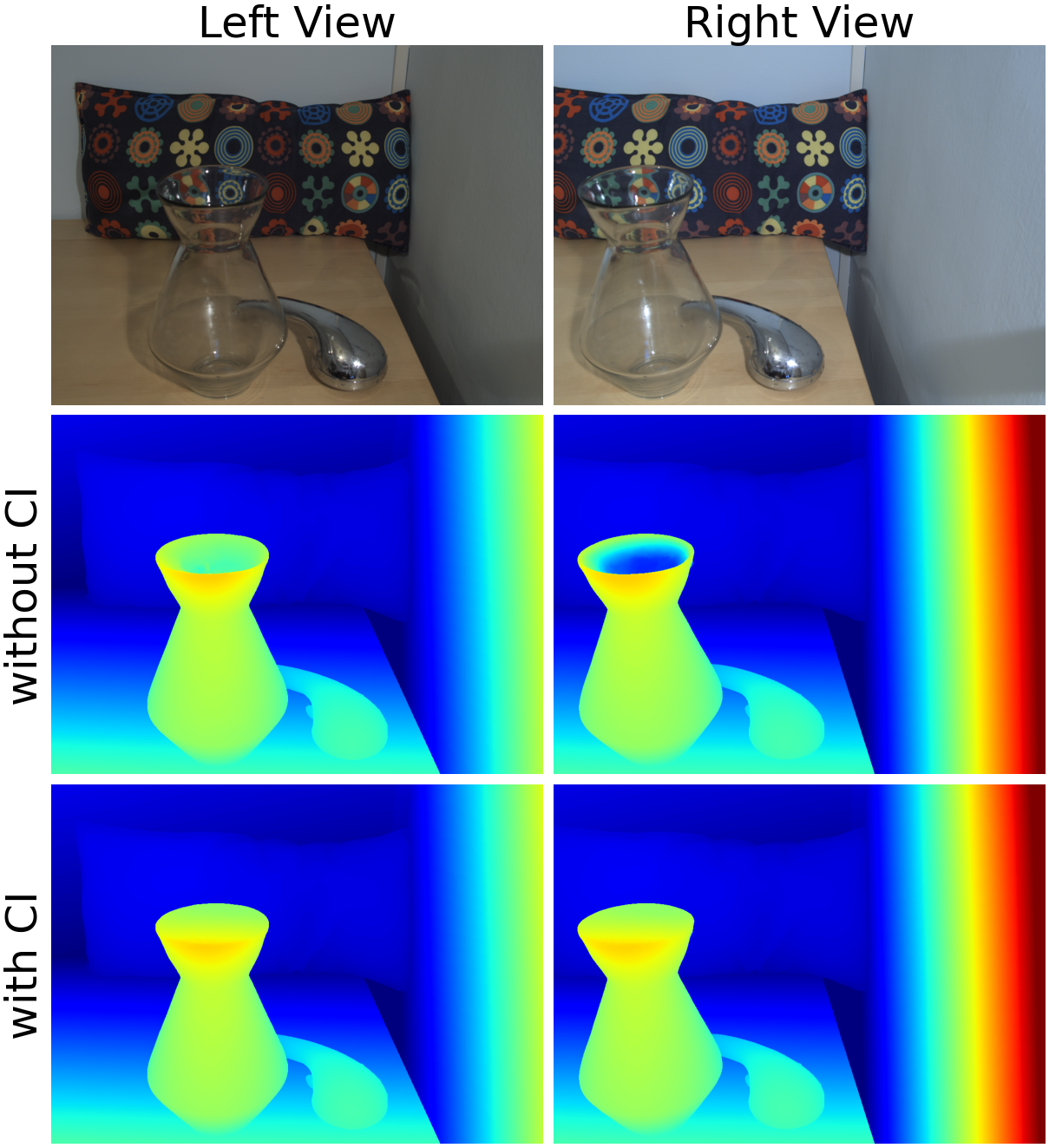}
          \caption{Pillow\_im5}
          \label{fig:booster-a}
      \end{subfigure}
      \hfill
        \begin{subfigure}[b]{0.245\textwidth}
          \centering
          \includegraphics[width=\textwidth]{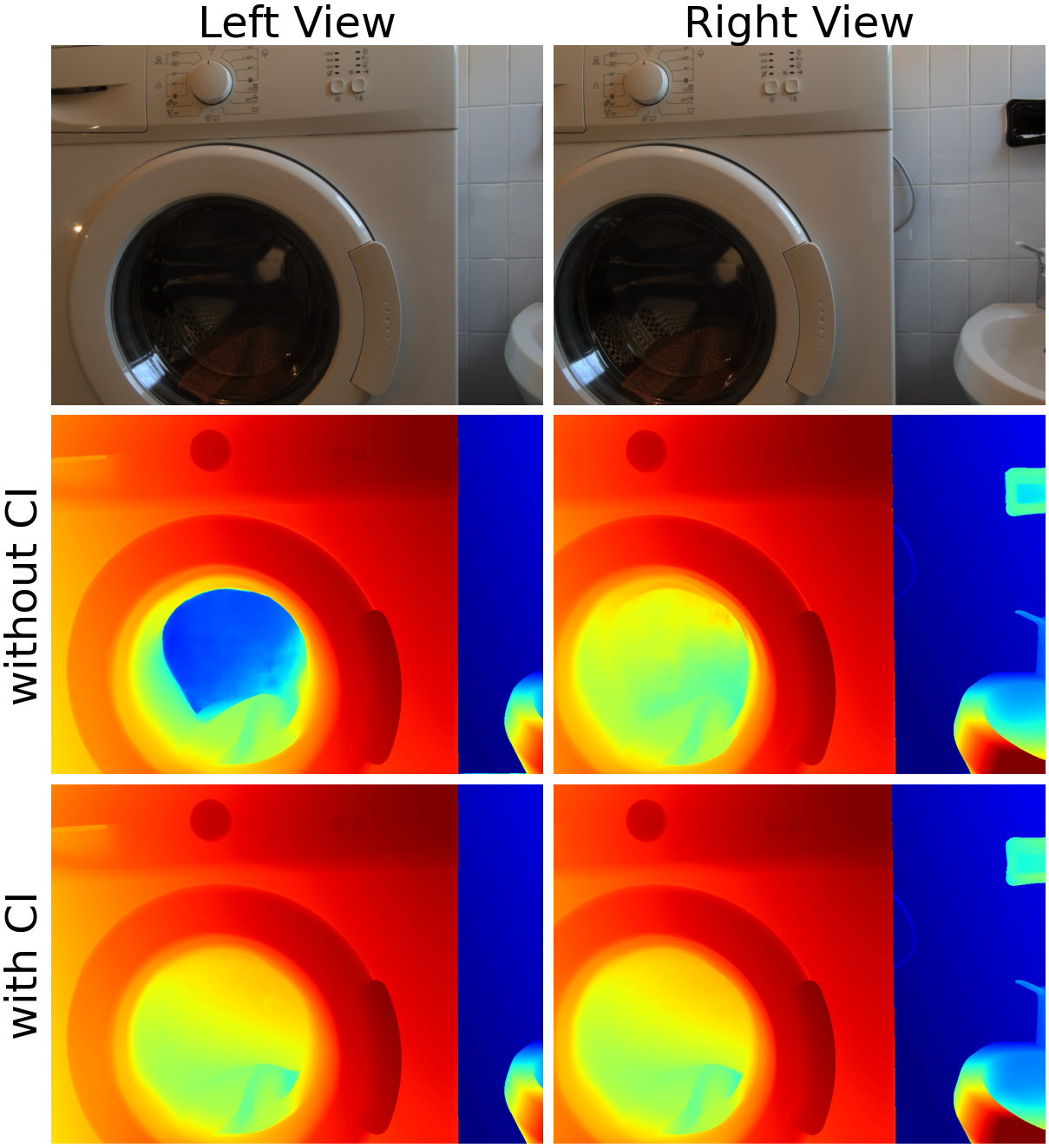}
          \caption{Washer1\_im1}
          \label{fig:booster-d}
      \end{subfigure}
      \hfill
      \begin{subfigure}[b]{0.245\textwidth}
          \centering
          \includegraphics[width=\textwidth]{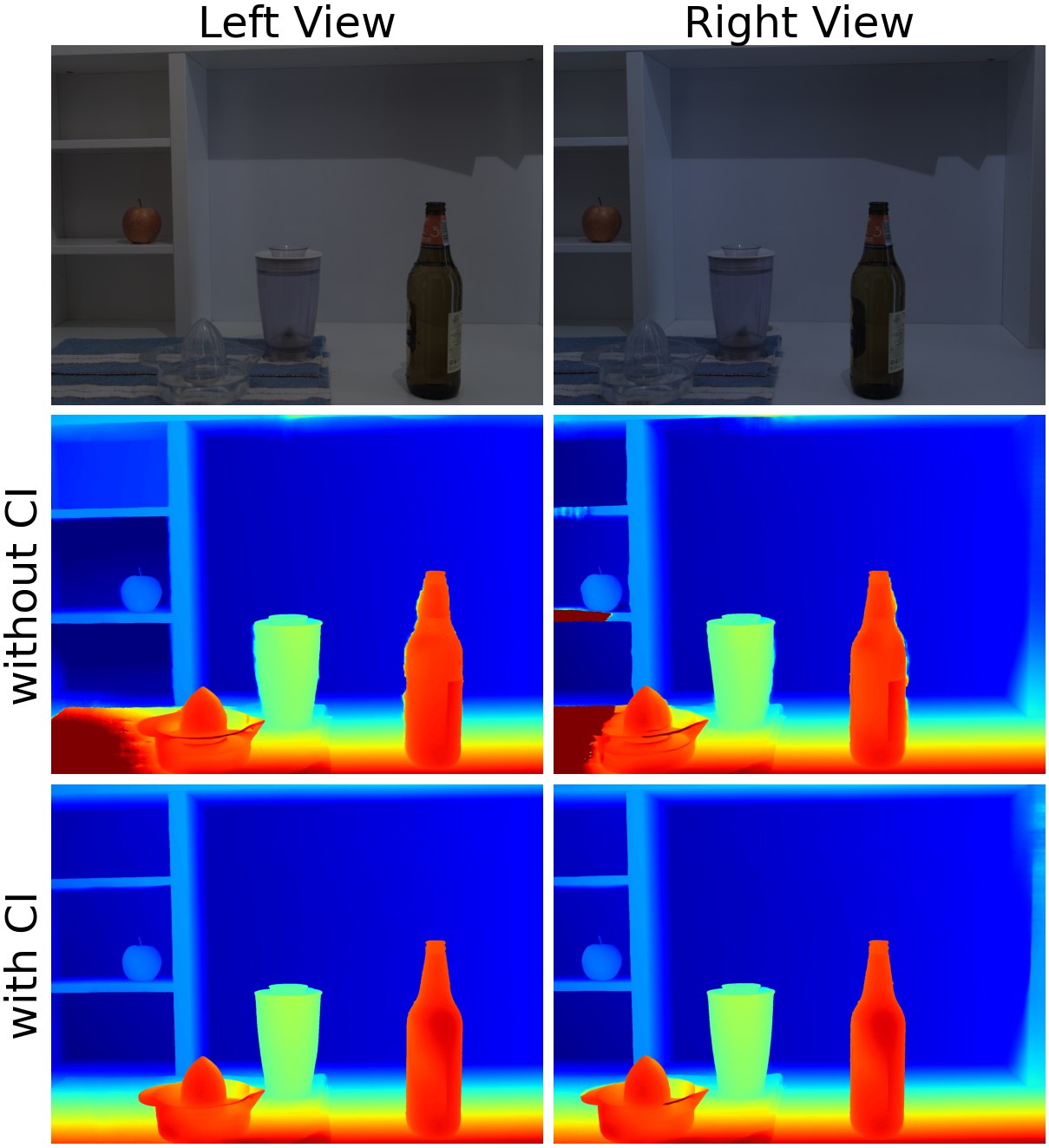}
          \caption{Blender\_im5}
          \label{fig:booster-b}
      \end{subfigure}
      \caption{FoundationStereo zero-shot predictions on selected \textbf{Booster}~\cite{booster} (half resolution) samples with and without our cross-injection method. CI corrects errors caused by reflections (a), transparency (b+c) and other random failure modes (d).}
      \label{fig:booster-samples}
  \end{figure*}

\begin{figure*}[p]
      \centering
      \begin{subfigure}[b]{0.49\textwidth}
          \centering
          \includegraphics[width=\textwidth]{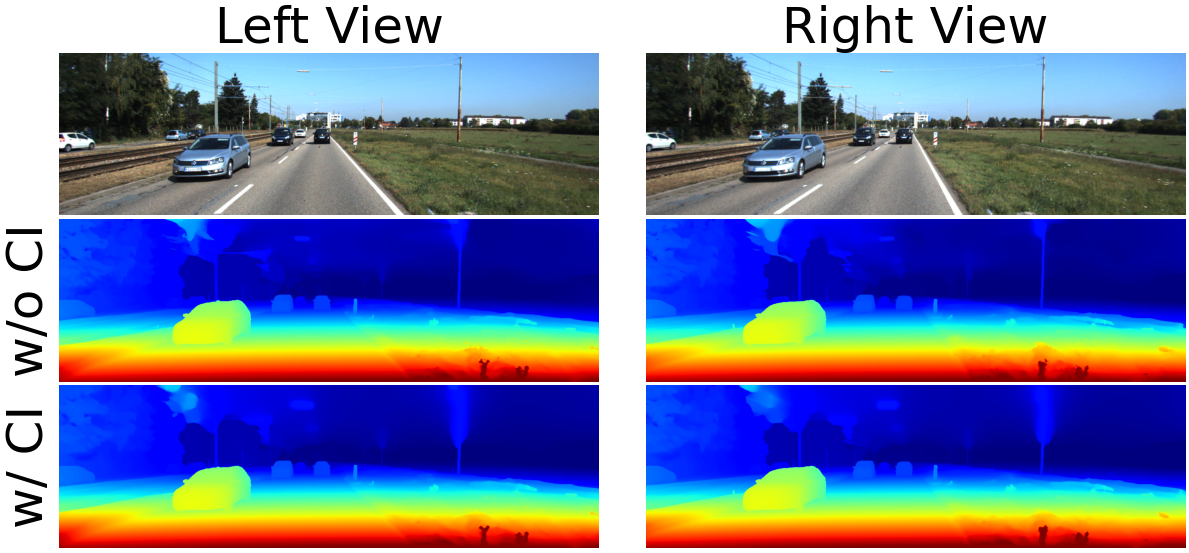}
          \caption{000041}
          \label{fig:kitti-a}
      \end{subfigure}
      \hfill
      \begin{subfigure}[b]{0.49\textwidth}
          \centering
          \includegraphics[width=\textwidth]{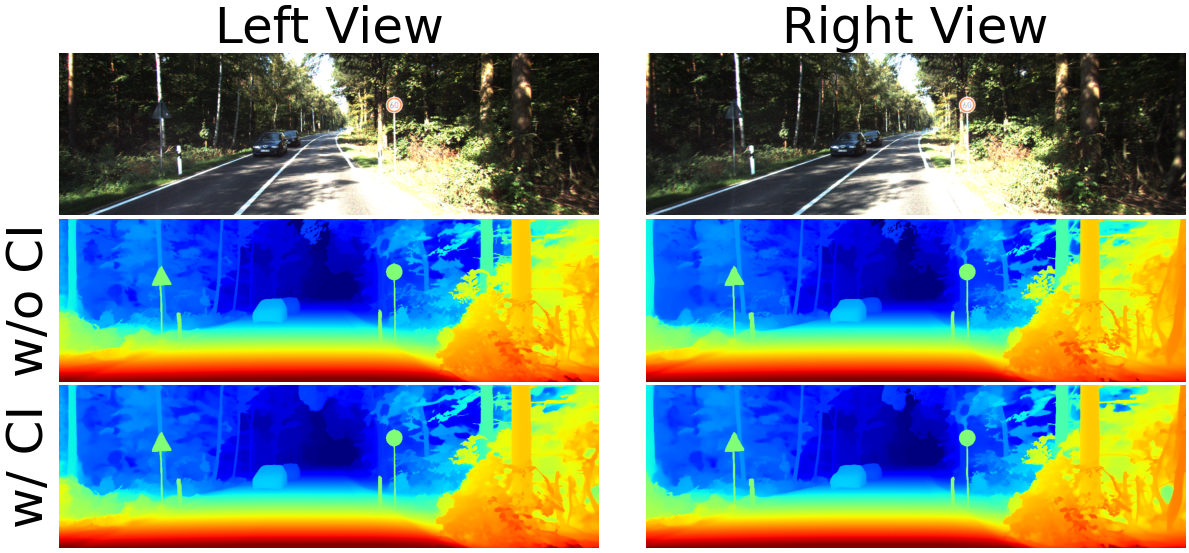}
          \caption{000064}
          \label{fig:kitti-b}
      \end{subfigure}
      \caption{FoundationStereo zero-shot predictions on selected \textbf{KITTI-15}~\cite{kitti15} samples with and without our cross-injection method.
      (a) shows occasional ``smearing'' effects on thin vertical structures such as traffic signs or street lamps in front of the sky. Interestingly, this failure mode was already present in FoundationStereo, but our method seems to amplify it. (b) shows an example where both models predict equally sharp edges on road signs and high frequency details (vegetation).
      }
      \label{fig:kitti-samples}
  \end{figure*}

In Figs.~\ref{fig:eth3d-samples}, \ref{fig:booster-samples} and \ref{fig:kitti-samples}, we show qualitative samples for three of the benchmark datasets used. We selected samples where CI makes a visible difference compared to standard FoundationStereo. Clear consistency improvements are evident for transparent and reflective surfaces (Figs.~\ref{fig:eth3d-a}, \ref{fig:booster-c}, \ref{fig:booster-d}). Interestingly, when resolving ambiguity the model commits to the first physical layer most of the time, even though this behavior is not enforced by the training data. CI also resolves other failures that occur in independent single-view passes, e.g.\ challenging cases caused by lens flare (Fig.~\ref{fig:eth3d-b}), reflections (Fig.~\ref{fig:eth3d-c}), or textureless regions (Fig.~\ref{fig:booster-a}).

We also do not wish to conceal the drawbacks of our method: for example, one can observe a slightly reduced level of detail in high-frequency signal (the background shrubs in Fig.~\ref{fig:eth3d-b}). We further observed an occasional failure mode for FoundationStereo + CI on KITTI-15, where thin vertical structures such as street lamps are ``smeared'' when located in front of the sky. This pattern already occurs for standard FoundationStereo and cannot be reproduced by \SSMM-XL (neither the original nor the fine-tuned version). This leads us to conclude that CI is not the cause of the failure mode, but rather amplifies it.

\section{Discussion}
\label{sec:conclusion}

\paragraph{Why does coupling both views improve single-view accuracy?}
Beyond test-time agreement, the coupling appears to aid training in three ways:
First, it \emph{extends supervision}: many stereo datasets label only the left view, so the right pass is otherwise trained only indirectly; coupling the two views turns single-view labels into a learning signal for both.
Second, it acts as a \emph{geometric regularizer}: requiring the two per-view disparities to agree constrains otherwise under-determined predictions.
Appendix~\ref{sec:regions} shows the constraint acting where it should: the change concentrates where the two views disagree and vanishes where they already agree, offset by a mild regression at occluded pixels, where the constraint does not apply.
Third, the coupling is \emph{soft}: the signed inconsistency is provided as an input cue rather than a hard constraint, allowing the model to rely on it where it is trustworthy and ignore it where it is not. More broadly, this suggests that supervising only the left view may be the wrong default even when only left-view disparity is sought: future stereo models may benefit from supervising both views in one form or another, whether through coupled per-view passes (such as cross-injection) or a fully symmetric design.

\paragraph{Limitations.}
CI assumes a RAFT-style iterative refinement stage; it does not apply to single-shot cost-volume regressors, nor to models built on transformer attention or cross-view completion without a recurrent refiner, and future architectures may move away from the iterative paradigm.
We study CI mainly in a fine-tuning setting. It also works when training from scratch (Appendix~\ref{sec:scratch}), but its disparity error converges more slowly at first: the consistency term is unreliable early in training, when both views are still far from correct. A ramp-up or a trust-weighted consistency loss~\cite{tonioni2017adaptation} may close this gap, which we leave to future work.
Lastly, CI needs two passes and recomputes the warp and co-visibility mask at every step, so its runtime grows with the iteration count (Appendix~\ref{sec:cost}).

\paragraph{Outlook.}
A natural next step is to push consistency upstream of the refiner. Cross-injection couples the two passes only during iterative refinement; the initial disparity that seeds each pass is still decoded independently from its own matching volume, leaving the refiner to reconcile two starting points that need not agree. Making that initialization left-right consistent would give both passes a coherent starting point and could compound the gains we report. Some architectures are especially well-suited to this: \SSMM's optimal-transport matching yields a (near) doubly stochastic assignment. We note that the transpose of such an assignment already encodes the reverse, right-to-left correspondence, so both views' initial disparities can in principle be read from a single volume. The open challenge is to obtain such symmetric initial predictions without sacrificing per-view accuracy to the added constraint; combined with cross-injection in the refinement stage, this points toward stereo matchers that are left-right consistent end to end.

\clearpage
{
    \small
    \bibliographystyle{ieeenat_fullname}
    \bibliography{references}

@STRING{cvpr = "CVPR"}

@STRING{cvprw = "CVPR Workshops"}

@STRING{iccv = "ICCV"}

@STRING{eccv = "ECCV"}

@STRING{ijcv = "IJCV"}

@STRING{pami = "IEEE TPAMI"}

@STRING{threedv = "3DV"}

@STRING{bmvc = "BMVC"}

@STRING{gcpr = "GCPR"}

@STRING{aaai = "AAAI"}

@STRING{neurips = "NeurIPS"}

@STRING{jmlr = "JMLR"}

@STRING{ral = "IEEE RA-L"}

@inproceedings{wen2025foundationstereo,
  title = {{FoundationStereo}: Zero-Shot Stereo Matching},
  author={Wen, Bowen and Trepte, Matthew and Aribido, Joseph and Kautz, Jan and Gallo, Orazio and Birchfield, Stan},
  booktitle=cvpr,
  year={2025}
}

@inproceedings{min2025s2m2,
  title={{S2M2}: Scalable Stereo Matching Model for Reliable Depth Estimation},
  author={Min, Junhong and Jeon, Youngpil and Kim, Jimin and Choi, Minyong},
  booktitle=iccv,
  xpages={26729--26739},
  year={2025}
}

@inproceedings{lipson2021raftstereo,
  title={{RAFT-Stereo}: Multilevel Recurrent Field Transforms for Stereo Matching},
  author={Lipson, Lahav and Teed, Zachary and Deng, Jia},
  booktitle=threedv,
  year={2021}
}

@article{marr1976cooperative,
  title={Cooperative Computation of Stereo Disparity},
  author={Marr, David and Poggio, Tomaso},
  journal={Science},
  xvolume={194},
  xnumber={4262},
  xpages={283--287},
  year={1976}
}

@article{belhumeur1996bayesian,
  title={A {Bayesian} approach to binocular stereopsis},
  author={Belhumeur, Peter},
  journal=ijcv,
  xvolume={19},
  xnumber={3},
  xpages={237--260},
  year={1996}
}

@inproceedings{middlebury,
  title={High-resolution stereo datasets with subpixel-accurate ground truth},
  author={Scharstein, Daniel and Hirschm{\"u}ller, Heiko and Kitajima, York and Krathwohl, Greg and Ne{\v{s}}i{\'c}, Nera and Wang, Xi and Westling, Porter},
  booktitle=gcpr,
  year={2014}
}

@inproceedings{eth3d,
  title={A multi-view stereo benchmark with high-resolution images and multi-camera videos},
  author={Sch\"{o}ps, Thomas and Sch\"{o}nberger, Johannes and Galliani, Silvano and Sattler, Torsten and Schindler, Konrad and Pollefeys, Marc and Geiger, Andreas},
  booktitle=cvpr,
  xpages={3260--3269},
  year={2017}
}

@article{booster,
  title={Booster: a benchmark for depth from images of specular and transparent surfaces},
  author={Ramirez, Pierluigi Zama and Costanzino, Alex and Tosi, Fabio and Poggi, Matteo and Salti, Samuele and Mattoccia, Stefano and Di Stefano, Luigi},
  journal=pami,
  xvolume={46},
  xnumber={1},
  xpages={85--102},
  year={2024}
}

@inproceedings{kitti12,
  title={Are We Ready for Autonomous Driving? The {KITTI} Vision Benchmark Suite},
  author={Geiger, Andreas and Lenz, Philip and Urtasun, Raquel},
  booktitle=cvpr,
  year={2012}
}

@inproceedings{kitti15,
  title={Object scene flow for autonomous vehicles},
  author={Menze, Moritz and Geiger, Andreas},
  booktitle=cvpr,
  xpages={3061--3070},
  year={2015}
}

@inproceedings{sceneflow,
  title={A large dataset to train convolutional networks for disparity, optical flow, and scene flow estimation},
  author={Mayer, Nikolaus and Ilg, Eddy and Hausser, Philip and Fischer, Philipp and Cremers, Daniel and Dosovitskiy, Alexey and Brox, Thomas},
  booktitle=cvpr,
  year={2016}
}

@article{vkitti2,
  title={Virtual {KITTI} 2},
  author={Cabon, Yohann and Murray, Naila and Humenberger, Martin},
  journal={arXiv:2001.10773},
  year={2020}
}

@inproceedings{fallingthings,
  title={{Falling Things}: A synthetic dataset for {3D} object detection and pose estimation},
  author={Tremblay, Jonathan and To, Thang and Birchfield, Stan},
  booktitle=cvprw,
  year={2018}
}

@inproceedings{sintel,
  title={A naturalistic open source movie for optical flow evaluation},
  author={Butler, Daniel and Wulff, Jonas and Stanley, Garrett and Black, Michael J.},
  booktitle=eccv,
  year={2012}
}

@inproceedings{li2022crestereo,
  title={Practical stereo matching via cascaded recurrent network with adaptive correlation},
  author={Li, Jiankun and Wang, Peisen and Xiong, Pengfei and Cai, Tao and Yan, Ziwei and Yang, Lei and Liu, Jiangyu and Fan, Haoqiang and Liu, Shuaicheng},
  booktitle=cvpr,
  year={2022}
}

@article{instereo2k,
  title={{InStereo2k}: A large real dataset for stereo matching in indoor scenes},
  author={Bao, Wei and Wang, Wei and Xu, Yuhua and Guo, Yulan and Hong, Siyu and Zhang, Xiaohu},
  journal={Science China Information Sciences},
  year={2020}
}

@inproceedings{teed2020raft,
  author    = {Teed, Zachary and Deng, Jia},
  title     = {{RAFT}: Recurrent All-Pairs Field Transforms for Optical Flow},
  booktitle = eccv,
  year      = {2020}
}

@inproceedings{xu2023igev,
  author    = {Xu, Gangwei and Wang, Xianqi and Ding, Xiaohuan and Yang, Xin},
  title     = {Iterative Geometry Encoding Volume for Stereo Matching},
  booktitle = cvpr,
  year      = {2023}
}

@inproceedings{wang2024selective,
  author    = {Wang, Xianqi and Xu, Gangwei and Jia, Hao and Yang, Xin},
  title     = {{Selective-Stereo}: Adaptive Frequency Information Selection for Stereo Matching},
  booktitle = cvpr,
  year      = {2024}
}

@inproceedings{jiang2025defom,
  author    = {Jiang, Hualie and Lou, Zhiqiang and Ding, Laiyan and Xu, Rui and Tan, Minglang and Jiang, Wenjie and Huang, Rui},
  title     = {{DEFOM-Stereo}: Depth Foundation Model Based Stereo Matching},
  booktitle = cvpr,
  year      = {2025}
}

@inproceedings{kendall2017gcnet,
  author    = {Kendall, Alex and Martirosyan, Hayk and Dasgupta, Saumitro and Henry, Peter and Kennedy, Ryan and Bachrach, Abraham and Bry, Adam},
  title     = {End-to-End Learning of Geometry and Context for Deep Stereo Regression},
  booktitle = iccv,
  year      = {2017}
}

@inproceedings{chang2018psmnet,
  author    = {Chang, Jia-Ren and Chen, Yong-Sheng},
  title     = {Pyramid Stereo Matching Network},
  booktitle = cvpr,
  year      = {2018}
}

@inproceedings{sun2005symmetric,
  author    = {Sun, Jian and Li, Yin and Kang, Sing Bing and Shum, Heung-Yeung},
  title     = {Symmetric Stereo Matching for Occlusion Handling},
  booktitle = cvpr,
  year      = {2005}
}

@inproceedings{li2018symmnet,
  author    = {Li, Ang and Yuan, Zejian},
  title     = {{SymmNet}: A Symmetric Convolutional Neural Network for Occlusion Detection},
  booktitle = bmvc,
  year      = {2018}
}

@inproceedings{li2021sttr,
  author    = {Li, Zhaoshuo and Liu, Xingtong and Drenkow, Nathan and Ding, Andy and Creighton, Francis and Taylor, Russell and Unberath, Mathias},
  title     = {Revisiting Stereo Depth Estimation From a Sequence-to-Sequence Perspective with Transformers},
  booktitle = iccv,
  year      = {2021}
}

@article{xu2023unimatch,
  author  = {Xu, Haofei and Zhang, Jing and Cai, Jianfei and Rezatofighi, Hamid and Yu, Fisher and Tao, Dacheng and Geiger, Andreas},
  title   = {Unifying Flow, Stereo and Depth Estimation},
  journal = pami,
  year    = {2023}
}

@article{wang2022pasmnet,
  title={Parallax attention for unsupervised stereo correspondence learning},
  author={Wang, Longguang and Guo, Yulan and Wang, Yingqian and Liang, Zhengfa and Lin, Zaiping and Yang, Jungang and An, Wei},
  journal=pami,
  xvolume={44},
  xnumber={4},
  xpages={2108--2125},
  year={2022}
}

@inproceedings{godard2017monodepth,
  author    = {Godard, Cl\'ement and Mac Aodha, Oisin and Brostow, Gabriel},
  title     = {Unsupervised Monocular Depth Estimation with Left-Right Consistency},
  booktitle = cvpr,
  year      = {2017}
}

@inproceedings{garg2016unsupervised,
  author    = {Garg, Ravi and Kumar BG, Vijay and Carneiro, Gustavo and Reid, Ian},
  title     = {Unsupervised {CNN} for Single View Depth Estimation: Geometry to the Rescue},
  booktitle = eccv,
  year      = {2016}
}

@inproceedings{tosi2019monoresmatch,
  author    = {Tosi, Fabio and Aleotti, Filippo and Poggi, Matteo and Mattoccia, Stefano},
  title     = {Learning Monocular Depth Estimation Infusing Traditional Stereo Knowledge},
  booktitle = cvpr,
  year      = {2019}
}

@inproceedings{poggi2018trinocular,
  author    = {Poggi, Matteo and Tosi, Fabio and Mattoccia, Stefano},
  title     = {Learning Monocular Depth Estimation with Unsupervised Trinocular Assumptions},
  booktitle = threedv,
  year      = {2018}
}

@inproceedings{pilzer2019refinedistill,
  author    = {Pilzer, Andrea and Lathuili\`ere, St\'ephane and Sebe, Nicu and Ricci, Elisa},
  title     = {Refine and Distill: Exploiting Cycle-Inconsistency and Knowledge Distillation for Unsupervised Monocular Depth Estimation},
  booktitle = cvpr,
  year      = {2019}
}

@article{zhong2017sssmnet,
  author        = {Zhong, Yiran and Dai, Yuchao and Li, Hongdong},
  title         = {Self-Supervised Learning for Stereo Matching with Self-Improving Ability},
  journal        = {arXiv:1709.00930},
  year          = {2017}
}

@article{wang2021pvstereo,
  author  = {Wang, Hengli and Fan, Rui and Cai, Peide and Liu, Ming},
  title   = {{PVStereo}: Pyramid Voting Module for End-to-End Self-Supervised Stereo Matching},
  journal = ral,
  year    = {2021}
}

@inproceedings{tonioni2017adaptation,
  author    = {Tonioni, Alessio and Poggi, Matteo and Mattoccia, Stefano and Di Stefano, Luigi},
  title     = {Unsupervised Adaptation for Deep Stereo},
  booktitle = iccv,
  year      = {2017}
}

@inproceedings{tonioni2019madnet,
  author    = {Tonioni, Alessio and Tosi, Fabio and Poggi, Matteo and Mattoccia, Stefano and Di Stefano, Luigi},
  title     = {Real-Time Self-Adaptive Deep Stereo},
  booktitle = cvpr,
  year      = {2019}
}

@inproceedings{meister2018unflow,
  author    = {Meister, Simon and Hur, Junhwa and Roth, Stefan},
  title     = {{UnFlow}: Unsupervised Learning of Optical Flow with a Bidirectional Census Loss},
  booktitle = aaai,
  year      = {2018}
}

@inproceedings{liu2019ddflow,
  author    = {Liu, Pengpeng and King, Irwin and Lyu, Michael and Xu, Jia},
  title     = {{DDFlow}: Learning Optical Flow with Unlabeled Data Distillation},
  booktitle = aaai,
  year      = {2019}
}

@inproceedings{liu2019selflow,
  author    = {Liu, Pengpeng and Lyu, Michael and King, Irwin and Xu, Jia},
  title     = {{SelFlow}: Self-Supervised Learning of Optical Flow},
  booktitle = cvpr,
  year      = {2019}
}

@article{zbontar2016mccnn,
  author  = {\v{Z}bontar, Jure and LeCun, Yann},
  title   = {Stereo Matching by Training a Convolutional Neural Network to Compare Image Patches},
  journal = jmlr,
  xvolume  = {17},
  xnumber  = {65},
  xpages   = {1--32},
  year    = {2016}
}

@article{hirschmuller2008sgm,
  author  = {Hirschm\"{u}ller, Heiko},
  title   = {Stereo Processing by Semiglobal Matching and Mutual Information},
  journal = pami,
  xvolume  = {30},
  xnumber  = {2},
  xpages   = {328--341},
  year    = {2008}
}

@article{scharstein2002taxonomy,
  author  = {Scharstein, Daniel and Szeliski, Richard},
  title   = {A Taxonomy and Evaluation of Dense Two-Frame Stereo Correspondence Algorithms},
  journal = ijcv,
  xvolume  = {47},
  xnumber  = {1--3},
  xpages   = {7--42},
  year    = {2002}
}

@article{egnal2002halfocclusion,
  author  = {Egnal, Geoffrey and Wildes, Richard},
  title   = {Detecting Binocular Half-Occlusions: Empirical Comparisons of Five Approaches},
  journal = pami,
  xvolume  = {24},
  xnumber  = {8},
  xpages   = {1127--1133},
  year    = {2002}
}

@inproceedings{liang2018featureconstancy,
  author    = {Liang, Zhengfa and Feng, Yiliu and Guo, Yulan and Liu, Hengzhu and Chen, Wei and Qiao, Linbo and Zhou, Li and Zhang, Jianfeng},
  title     = {Learning for Disparity Estimation Through Feature Constancy},
  booktitle = cvpr,
  year      = {2018}
}

@inproceedings{chabra2019stereodrnet,
  author    = {Chabra, Rohan and Straub, Julian and Sweeney, Christopher and Newcombe, Richard and Fuchs, Henry},
  title     = {{StereoDRNet}: Dilated Residual {StereoNet}},
  booktitle = cvpr,
  year      = {2019}
}

@inproceedings{xu2022acvnet,
  author    = {Xu, Gangwei and Cheng, Junda and Guo, Peng and Yang, Xin},
  title     = {Attention Concatenation Volume for Accurate and Efficient Stereo Matching},
  booktitle = cvpr,
  year      = {2022}
}

@inproceedings{weinzaepfel2022croco,
  author    = {Weinzaepfel, Philippe and Leroy, Vincent and Lucas, Thomas and Br\'egier, Romain and Cabon, Yohann and Arora, Vaibhav and Antsfeld, Leonid and Chidlovskii, Boris and Csurka, Gabriela and Revaud, J\'er\^ome},
  title     = {{CroCo}: Self-Supervised Pre-training for {3D} Vision Tasks by Cross-View Completion},
  booktitle = neurips,
  year      = {2022}
}

@inproceedings{weinzaepfel2023crocov2,
  author    = {Weinzaepfel, Philippe and Lucas, Thomas and Leroy, Vincent and Cabon, Yohann and Arora, Vaibhav and Br\'egier, Romain and Csurka, Gabriela and Antsfeld, Leonid and Chidlovskii, Boris and Revaud, J\'er\^ome},
  title     = {{CroCo} v2: Improved Cross-View Completion Pre-training for Stereo Matching and Optical Flow},
  booktitle = iccv,
  year      = {2023}
}

@inproceedings{gong2024crossview,
  author    = {Gong, Rui and Liu, Weide and Gu, Zaiwang and Yang, Xulei and Cheng, Jun},
  title     = {Learning Intra-view and Cross-view Geometric Knowledge for Stereo Matching},
  booktitle = cvpr,
  year      = {2024}
}

@inproceedings{jie2018lrcr,
  author    = {Jie, Zequn and Wang, Pengfei and Ling, Yonggen and Zhao, Bo and Wei, Yunchao and Feng, Jiashi and Liu, Wei},
  title     = {Left-Right Comparative Recurrent Model for Stereo Matching},
  booktitle = cvpr,
  year      = {2018}
}

@inproceedings{guo2018distillstereo,
  author    = {Guo, Xiaoyang and Li, Hongsheng and Yi, Shuai and Ren, Jimmy and Wang, Xiaogang},
  title     = {Learning Monocular Depth by Distilling Cross-Domain Stereo Networks},
  booktitle = eccv,
  year      = {2018}
}

@inproceedings{elchemaly2025geometrypassthrough,
  author    = {El Chemaly, Trishia and Goyal, Mohit and Duan, Tinglin and Phadnis, Vrushank and Khattar, Sakar and Vlaskamp, Bjorn and Kulshrestha, Achin and Turner, Eric Lee and Purohit, Aveek and Neiswander, Gregory and Tsotsos, Konstantine},
  title     = {Mind the {GAP}: Geometry Aware Passthrough Mitigates Cybersickness},
  booktitle = {CHI Extended Abstracts},
  year      = {2025}
}

@inproceedings{kim2021visualcomfortrl,
  author    = {Kim, Hak Gu and Park, Minho and Lee, Sangmin and Kim, Seongyeop and Ro, Yong Man},
  title     = {Visual Comfort Aware-Reinforcement Learning for Depth Adjustment of Stereoscopic {3D} Images},
  booktitle = aaai,
  year      = {2021}
}

@article{zhou2018stereomag,
  author  = {Zhou, Tinghui and Tucker, Richard and Flynn, John and Fyffe, Graham and Snavely, Noah},
  title   = {Stereo Magnification: Learning View Synthesis Using Multiplane Images},
  journal = {ACM TOG},
  year    = {2018}
}

@inproceedings{chen2021simsiam,
  author    = {Chen, Xinlei and He, Kaiming},
  title     = {Exploring Simple {Siamese} Representation Learning},
  booktitle = cvpr,
  year      = {2021}
}

@inproceedings{grill2020byol,
  author    = {Grill, Jean-Bastien and Strub, Florian and Altch\'e, Florent and Tallec, Corentin and Richemond, Pierre and Buchatskaya, Elena and Doersch, Carl and Pires, Bernardo Avila and Guo, Zhaohan Daniel and Azar, Mohammad Gheshlaghi and Piot, Bilal and Kavukcuoglu, Koray and Munos, R\'emi and Valko, Michal},
  title     = {Bootstrap Your Own Latent: A New Approach to Self-Supervised Learning},
  booktitle = neurips,
  year      = {2020}
}

@inproceedings{tarvainen2017meanteacher,
  author    = {Tarvainen, Antti and Valpola, Harri},
  title     = {Mean Teachers Are Better Role Models: Weight-Averaged Consistency Targets Improve Semi-Supervised Deep Learning Results},
  booktitle = neurips,
  year      = {2017}
}

@inproceedings{zhang2018dml,
  author    = {Zhang, Ying and Xiang, Tao and Hospedales, Timothy and Lu, Huchuan},
  title     = {Deep Mutual Learning},
  booktitle = cvpr,
  year      = {2018}
}

@inproceedings{yang2019drivingstereo,
  author    = {Yang, Guorun and Song, Xiao and Huang, Chaoqin and Deng, Zhidong and Shi, Jianping and Zhou, Bolei},
  title     = {{DrivingStereo}: A Large-Scale Dataset for Stereo Matching in Autonomous Driving Scenarios},
  booktitle = cvpr,
  year      = {2019}
}

@inproceedings{chen2024mocha,
  title={{MoCha-Stereo}: Motif Channel Attention Network for Stereo Matching},
  author={Chen, Ziyang and Long, Wei and Yao, He and Zhang, Yongjun and Wang, Bingshu and Qin, Yongbin and Wu, Jia},
  booktitle=cvpr,
  year={2024}
}

@inproceedings{bartolomei2025stereoanywhere,
  title={Stereo Anywhere: Robust Zero-Shot Deep Stereo Matching Even Where Either Stereo or Mono Fail},
  author={Bartolomei, Luca and Tosi, Fabio and Poggi, Matteo and Mattoccia, Stefano},
  booktitle=cvpr,
  year={2025}
}

@inproceedings{wang2018occaware,
  title={Occlusion Aware Unsupervised Learning of Optical Flow},
  author={Wang, Yang and Yang, Yi and Yang, Zhenheng and Zhao, Liang and Wang, Peng and Xu, Wei},
  booktitle=cvpr,
  year={2018}
}

@inproceedings{zhou2017unsupervisedstereo,
  title={Unsupervised Learning of Stereo Matching},
  author={Zhou, Chao and Zhang, Hong and Shen, Xiaoyong and Jia, Jiaya},
  booktitle=iccv,
  year={2017}
}

@inproceedings{loshchilov2019adamw,
  title={Decoupled weight decay regularization},
  author={Loshchilov, Ilya and Hutter, Frank},
  booktitle={ICLR},
  year={2019}
}

@article{chaurasia2020passthrough,
  author = {Chaurasia, Gaurav and Nieuwoudt, Arthur and Ichim, Alexandru-Eugen and Szeliski, Richard and Sorkine-Hornung, Alexander},
  title = {Passthrough+: Real-time Stereoscopic View Synthesis for Mobile Mixed Reality},
  xvolume = {3},
  xnumber = {1},
  journal = {Proc. ACM Comput. Graph. Interact. Tech.},
  year = {2020}
  }

@inproceedings{yang2024depthanythingv2,
  title     = {Depth Anything {V2}},
  author    = {Yang, Lihe and Kang, Bingyi and Huang, Zilong and Zhao, Zhen and Xu, Xiaogang and Feng, Jiashi and Zhao, Hengshuang},
  booktitle = neurips,
  year      = {2024}
}
}
\clearpage
\begin{appendices}
\onecolumn
\setcounter{figure}{0}
\setcounter{table}{0}
\setcounter{section}{0}
\renewcommand{\thesection}{\Alph{section}}%
\renewcommand\thefigure{S.\arabic{figure}}
\renewcommand\thetable{S.\arabic{table}}
\begin{center}
    \begin{minipage}{\textwidth}
        \centering
        \fontsize{16}{20}\selectfont
        \textbf{Supplementary Materials}
    \end{minipage}
\end{center}

\vspace{12pt}
\section{Training Details}
\label{sec:training_details}

Cross-injection is applied via fine-tuning: we start from each model's publicly released left-only checkpoint, attach the cross-injection refiner with the injection pathway zero-initialized (an exact no-op at initialization, so step~0 reproduces the base model), and fine-tune the full network. All runs use AdamW~\cite{loshchilov2019adamw} with $\beta=(0.9,0.999)$ and no weight decay, gradient checkpointing (required by the two weight-shared per-view passes and the joint refinement loop), and $320\times736$ random crops. The disparity term is each base model's native sequence loss; to it we add the left-right consistency loss (coefficient $0.5$) with the bidirectional gradient routing of Sec.~\ref{sec:methods} and no trust weighting, and for the \SSMM models the base model's occlusion, confidence, and matching-cost terms are kept active. Tab.~\ref{tab:training_details} lists the per-model settings.

\paragraph{Update order.} The two passes can be coupled in two ways at each refinement step: (a) a Jacobi scheme, which computes both inconsistencies from the previous step's disparities and updates both passes simultaneously (as illustrated in Fig.~\ref{fig:arch}); or (b) a Gauss-Seidel scheme, which updates the left pass first and computes the right pass's inconsistency from the freshly updated left disparity. We ablated this using FoundationStereo, and the two schemes performed indistinguishably: all disparity and LR-BP-1 columns of Tab.~\ref{tab:ablation} agreed within one checkpoint standard deviation. Therefore, all reported results in this paper use the Jacobi scheme. It avoids recomputing the right view's cue within a step and allows both updates to run in parallel, making it slightly more efficient.

\paragraph{Matched fine-tune controls.} Each ``+\,CI'' row of Tab.~\ref{tab:main_metrics} is paired with a control trained under the \emph{identical} recipe (same starting checkpoint, learning rate, schedule, data mixture, augmentation, batch size and number of updates), with only the cross-injection refiner and the left-right consistency loss removed. The control still predicts both views with two weight-shared passes and still receives the random flip-and-swap augmentation (probability $0.5$), so the sole difference is the coupling between the two passes. This is what allows a gain to be attributed to cross-injection rather than to the additional fine-tuning: the public checkpoints were trained on different mixtures, so ``fine-tuned beats not fine-tuned'' would otherwise be the confound. For FoundationStereo the control is exactly row~(a) of the ablation in Tab.~\ref{tab:ablation}; for \SSMM\ it is a dedicated run per model size.

\paragraph{Training data.} The FoundationStereo fine-tune and the ablation of Sec.~\ref{sec:ablation} use a subset of FoundationStereo's original training mixture: the FoundationStereo Dataset~\cite{wen2025foundationstereo}, SceneFlow~\cite{sceneflow}, FallingThings~\cite{fallingthings}, Virtual\,KITTI\,2~\cite{vkitti2}, and Sintel~\cite{sintel}, omitting CREStereo~\cite{li2022crestereo} and InStereo2K~\cite{instereo2k}. The \SSMM\ fine-tunes use a subset of \SSMM's original mixture (the FoundationStereo Dataset~\cite{wen2025foundationstereo}, SceneFlow~\cite{sceneflow}, FallingThings~\cite{fallingthings}, Virtual\,KITTI\,2~\cite{vkitti2}, and DrivingStereo~\cite{yang2019drivingstereo}), introducing no data the base model was not already trained on. None of the five evaluation benchmarks (Middlebury, ETH3D, KITTI-12, KITTI-15, Booster) appears in any training mixture.
The runs differ in scale, precision, and schedule. FoundationStereo is fine-tuned in \texttt{bf16} at an effective batch of $128$ ($32$ H100 GPUs) for $50$k steps, with a step-drop from $10^{-5}$ to $5\times10^{-6}$ over the final $10$k. Both \SSMM\ models train in \texttt{fp32}, also at an effective batch of $128$ on $32$ GPUs, for $4000$ optimizer updates with a cosine schedule at a peak learning rate of $2\times10^{-6}$. \SSMM-XL halves the per-GPU batch and compensates with two gradient-accumulation steps, so all \SSMM\ arms see the same effective batch and the same number of updates.

\paragraph{Reporting point.} For every \SSMM\ row we report the \emph{final} step, for both the control and the cross-injection arm. This is deliberate: the cosine is defined over exactly the number of updates the run performs, so the learning rate anneals and the run ends at its operating point rather than being truncated mid-flight. No snapshot is selected on the evaluation benchmarks for either arm. The single free choice is the schedule \emph{length}, fixed once from a preliminary longer run and then applied identically to all four \SSMM\ arms; because it is shared, it cannot favor cross-injection over its control. We verified that the comparison does not depend on this choice: cross-injection improves left-right consistency over the matched control at \emph{every} evaluated step of both runs ($15/15$ for \SSMM-L, $12/12$ for \SSMM-XL), and improves at least four of the five disparity columns at all but two of those steps.

\begin{table}[ht!]
  \centering
  \caption{Fine-tuning configuration for the models in Tab.~\ref{tab:main_metrics}. Each column applies to \emph{both} the cross-injection run and its matched no-CI control, which differ only in the cross-injection refiner and the left-right consistency loss. All runs start from the public left-only checkpoint with the cross-injection pathway zero-initialized and fine-tune the full network on $320\times736$ crops. The \SSMM\ cosine is defined over exactly the number of updates performed, so the learning rate anneals and the final step is the reported one.}
  \label{tab:training_details}
  \begin{tabular}{lccc}
    \toprule
     & FS & \SSMM-L & \SSMM-XL \\
    \midrule
    Optimizer              & \multicolumn{3}{c}{AdamW, $\beta=(0.9,0.999)$, wd $=0$} \\
    Peak LR                & $10^{-5}$ & $2\times10^{-6}$ & $2\times10^{-6}$ \\
    LR schedule            & step-drop & cosine & cosine \\
    Optimizer updates      & $50$k & $4$k & $4$k \\
    Precision              & \texttt{bf16} & \texttt{fp32} & \texttt{fp32} \\
    GPUs ($\times$bs/GPU)  & $32\times4$ & $32\times4$ & $32\times2$ \\
    Grad.\ accumulation    & $1$ & $1$ & $2$ \\
    Effective batch        & $128$ & $128$ & $128$ \\
    \bottomrule
  \end{tabular}
\end{table}

\FloatBarrier
\section{Inference Time Breakdown}
\label{sec:cost}

Cross-injection's cost is proportional to the refinement budget, since the warp and the co-visibility mask are recomputed for both views at every step. Tab.~\ref{tab:inference_cost} measures it at $480\times640$ in \texttt{fp32} (the frozen monocular branch stays \texttt{fp16}), batch~1, on a single A100, from the released checkpoints. The co-visibility test gathers over an index set whose size depends on the predicted disparities, so we time a stereo pair the network can actually solve rather than noise.
Any method that predicts both views runs the network twice, so the reference point is the two-pass baseline, not a single pass. The second pass costs exactly what the first one does ($+100\%$), and cross-injection then adds $215$\,ms on top of $896$\,ms, $+23.9\%$; repeated measurements vary by about a point. Against a single left-only pass the full method is $2.5\times$, of which $2.0\times$ is the second pass that any both-view predictor requires.

\begin{table}[ht!]
\centering
\caption{\textbf{What cross-injection costs at inference, and what that is a fraction of.} Wall-clock at $480\times640$ in \texttt{fp32} (the frozen monocular branch stays \texttt{fp16}), batch~1, on one A100-SXM4-80GB; median of 36 CUDA-event-timed runs per configuration, the configurations interleaved round-robin so that GPU clock drift cannot be mistaken for overhead. \emph{1 pass} predicts left disparity only. \emph{2 passes} is the standard way to obtain both views (two weight-shared, uncoupled passes) and is the baseline cross-injection should be judged against, because any both-view predictor pays it: the second pass costs what the first one does ($+100\%$). Cross-injection is then charged per refinement step, which is why the same mechanism costs five to seven times less, relative to the two-pass baseline, on models that iterate three times instead of $32$.}
\label{tab:inference_cost}
\begin{tabular}{lrrrrr}
\toprule
Model & 1 pass & 2 passes & \;+\,CI & \multicolumn{2}{c}{CI overhead} \\
\cmidrule(lr){5-6}
 & (ms) & (ms) & (ms) & vs.\ 2 passes & vs.\ 1 pass \\
\midrule
FoundationStereo ($32$ iters) & 447 & 896 & 1111 & $+23.9\%$ & $+148\%$ \\
\SSMM-L ($3$ iters) & 141 & 283 & 298 & $+5.2\%$ & $+111\%$ \\
\SSMM-XL ($3$ iters) & 240 & 480 & 498 & $+3.6\%$ & $+107\%$ \\
\bottomrule
\end{tabular}
\end{table}

\paragraph{Inference time breakdown.} Refinement is $58\%$ of a FoundationStereo pass, and cross-injection adds $3.3$\,ms per iteration and view end to end, of which $0.33$\,ms is the warp, $1.60$\,ms the co-visibility mask, and $1.31$\,ms the widened GRU input, compared with $7.0$\,ms for one GRU update (Tab.~\ref{tab:inference_stages}). The mask, not the warp, is what the cue costs, because the uniqueness test scatters over a data-dependent index set. Dropping the mask channel (Appendix~\ref{sec:lrcr}) would not remove this cost, since the mask also zeroes the residual at non-co-visible pixels. Because the charge is per step, the same mechanism costs roughly five to seven times less on \SSMM, which runs three iterations instead of $32$: $+5.2\%$ for L and $+3.6\%$ for XL, against $+23.9\%$.

\paragraph{Latency-accuracy trade-off.} The iteration count $K$ is a free knob at inference, so the question the overhead raises is which model is better at equal latency. Cross-injection at $K{=}16$ is better than the baseline at $K{=}32$ on disparity error \emph{and} on left-right consistency, on both benchmarks, at $84\%$ of its runtime ($739$ against $882$\,ms); for consistency alone, $K{=}4$ already suffices, at $56\%$ of the runtime (Tab.~\ref{tab:iters_pareto}). The per-step overhead is more than repaid by needing fewer steps. On ETH3D, the uncoupled baseline even drifts apart as it iterates (LR-BP-1 $4.98\rightarrow5.59$ from $K{=}4$ to $32$), while cross-injection keeps converging ($3.46\rightarrow3.24$). On every column the two tables share, the baseline's $K{=}32$ row reproduces Tab.~\ref{tab:main_metrics} exactly; the CI row differs by at most $0.03$ because it is evaluated in \texttt{fp32} here rather than \texttt{bf16}.

\paragraph{Cost-volume symmetry.} In the FoundationStereo architecture, part of the second pass can be avoided outright. The right view's group-wise correlation volume is exactly a gather from the left one, $\mathrm{cv}[d,w{+}d]$, which we verified reproduces the flipped construction bit-for-bit; taking it instead of rebuilding removes the second feature extraction and volume construction, roughly $82$ down to $4$\,ms, or $8.8\%$ of a two-view forward. Only the raw correlation transposes, however: the concatenation volume is rebuilt, and the 3D-UNet filtering, the context encoder and the whole refinement loop are conditioned on the view's own features and run twice either way. The filtering alone costs $92.5$\,ms per view (Tab.~\ref{tab:inference_stages}), more than the transpose saves. A design that also shared the aggregation, decoding both views from a single volume (Sec.~\ref{sec:conclusion}), could roughly double the saving, but no more: refinement is $58\%$ of a pass and each view must refine its own disparity field.

\begin{table}[ht!]
\centering
\caption{\textbf{Inference time breakdown.} Per-stage GPU time for FoundationStereo at $32$ refinement iterations, under the conditions of Tab.~\ref{tab:inference_cost}; CUDA events are recorded inline and resolved after a single synchronization, so no stage boundary stalls the stream. Columns are the three configurations of Tab.~\ref{tab:inference_cost}. Refinement accounts for most of a pass, which is why a per-step cost is material for this model. Per iteration and per view, the cue costs 0.33\,ms of warp and 1.60\,ms of co-visibility mask, against 7.0\,ms for one GRU update: the injected cue is a cheap tensor operation, not a second network, and the mask, not the warp, is what it costs. The sum of stages falls $0.8$--$1.6\%$ short of the independently measured end-to-end time, which bounds the unattributed time.}
\label{tab:inference_stages}
\footnotesize
\begin{tabular}{lrrr}
\toprule
Stage (ms) & 1 pass & 2 passes & \;+\,CI \\
\midrule
Feature extraction (both images) & 54.6 & 111.7 & 113.4 \\
\quad stem & 0.311 & 0.610 & 0.613 \\
Cost-volume construction & 22.4 & 48.3 & 49.3 \\
Cost-volume filtering (3D UNet) & 92.5 & 185.0 & 184.7 \\
Initial disparity decode & 0.125 & 0.250 & 0.251 \\
Geometry-encoding pyramid & 1.97 & 3.94 & 3.92 \\
\quad other (volume stage) & 0.009 & 0.018 & 0.017 \\
Context encoder & 11.0 & 22.1 & 22.0 \\
Refinement: geometry lookup & 16.4 & 33.0 & 32.1 \\
Refinement: GRU update & 222.5 & 447.6 & 531.6 \\
Refinement: upsampling head & 17.6 & 35.3 & 34.5 \\
\;+\,\textbf{CI: partner warp} & -- & -- & \textbf{21.3} \\
\;+\,\textbf{CI: co-visibility mask} & -- & -- & \textbf{102.6} \\
\quad other (refinement) & 0.308 & 0.635 & 4.83 \\
\midrule
Sum of stages & 439.8 & 888.4 & 1101.3 \\
End-to-end & 447.1 & 896.5 & 1110.6 \\
\bottomrule
\end{tabular}
\end{table}

\begin{table}[ht!]
\centering
\caption{\textbf{Accuracy against latency.} Latency is measured at $480\times640$ as in Tab.~\ref{tab:inference_cost}, accuracy at each benchmark's native resolution. \textbf{Cross-injection at $K{=}16$ is better than the baseline at $K{=}32$ on every column shown, at a lower latency}: the extra per-step cost is more than repaid by needing fewer steps. On the columns shared with Tab.~\ref{tab:main_metrics}, the $K{=}32$ rows reproduce it; the CI row differs by at most $0.03$ because it is evaluated in \texttt{fp32} here.}
\label{tab:iters_pareto}
\footnotesize
\begin{tabular}{lrrrrr}
\toprule
Model & Latency & \multicolumn{2}{c}{Middlebury-H} & \multicolumn{2}{c}{ETH3D} \\
 & (ms) & BP-2 & LR-BP-1 & BP-2 & LR-BP-1 \\
\midrule
FoundationStereo ($K=4$) & 469 & 3.43 & 10.54 & 0.50 & 4.98 \\
FoundationStereo ($K=8$) & 514 & 3.12 & 10.13 & 0.44 & 4.92 \\
FoundationStereo ($K=16$) & 635 & 2.91 & 10.13 & 0.44 & 5.49 \\
FoundationStereo ($K=24$) & 780 & 2.84 & 10.13 & 0.43 & 5.57 \\
FoundationStereo ($K=32$) & 882 & 2.77 & 9.89 & 0.43 & 5.59 \\
\midrule
\;+\,CI ($K=4$) & 494 & 3.15 & 9.54 & 0.41 & 3.46 \\
\;+\,CI ($K=8$) & 562 & 2.94 & 9.19 & 0.38 & 3.34 \\
\;+\,CI ($K=16$) & 739 & 2.72 & 9.05 & 0.37 & 3.28 \\
\;+\,CI ($K=24$) & 907 & 2.71 & 9.04 & 0.36 & 3.24 \\
\;+\,CI ($K=32$) & 1099 & 2.71 & 9.03 & 0.36 & 3.24 \\
\bottomrule
\end{tabular}
\end{table}

\clearpage
\section{Additional Cue Ablations}
\label{sec:lrcr}

In this section we take a closer look at \emph{which} consistency cue is injected. Our cue differs from an LRCR-style alternative in three choices, which we ablate separately (Tab.~\ref{tab:cue_ablation}): (1) signed vs.\ unsigned residual, (2) explicit residual vs.\ learned map, and (3) masked vs.\ unmasked. As a reference, ``no cue'' is the same recipe with the cross-injection refiner removed.
The learned variant follows the paradigm of LRCR~\cite{jie2018lrcr}, where a learned module infers the disagreement from the two disparity maps rather than subtracting them directly and injecting the residual.
Concretely, it stacks the view's own disparity and the warped partner disparity into two channels, normalizes both by the quarter-resolution disparity range, and maps them through three $3\times3$ convolutions ($32$ hidden channels, ReLU) and a sigmoid to a single attention channel in $[0,1]$.
It is \emph{not} the exact LRCR method, for which no public implementation is available to benchmark against, but an adaptation of its cue to our refiner-based injection, trained under our recipe with the consistency loss active so that the injected channel is the only thing that varies. Among other differences, LRCR routes the left-right consistency signal to the cost volume rather than the refiner stage, it uses no consistency loss term, and it reports no consistency metrics.
Tab.~\ref{tab:cue_ablation} shows that the sign is the choice that matters. Against the same recipe with nothing injected, the signed residual lowers LR-BP-1 by $17$--$46\%$ on ETH3D, KITTI-15 and Booster and by $3\%$ on Middlebury, whereas the unsigned residual helps substantially only on Booster ($11$--$12\%$; at most $4\%$ elsewhere) and the learned map changes consistency by at most about $3\%$. Without the sign, making the cue explicit helps little (unsigned vs.\ learned), and dropping the co-visibility mask changes LR-BP-1 by at most $0.09$ for the signed cue; both matter far less than the sign. On Middlebury, BP-2 stays within $2.61$--$2.71$, with the signed variants at the upper end; on ETH3D the signed variants are best ($0.58$--$0.59$ against $0.62$--$0.67$). In our setup (i.e., refiner-based injection), only the \emph{signed} residual substantially improves left-right consistency.

\begin{table*}[ht!]
\centering
\caption{\textbf{What the injected channel has to carry.} All variants share the recipe of Tab.~\ref{tab:ablation}(e), including the bidirectional gradient routing, and differ only in the content of the cross-injection channel, so each pair of rows that differs in one column isolates one factor. ``no cue'' is the same recipe with the cross-injection refiner removed, i.e., the consistency loss with nothing injected (Tab.~\ref{tab:ablation}(b) with bidirectional gradient routing; not a row of that table). ``learned'' is a learned attention map inspired by LRCR~\cite{jie2018lrcr}, adapted to our refiner-based injection (see text). \textbf{Finding:} only the \emph{signed} residual substantially improves left-right consistency over injecting nothing ($17$--$46\%$ lower LR-BP-1 on ETH3D, KITTI-15 and Booster, $3\%$ on Middlebury). The unsigned residual helps substantially only on Booster ($11$--$12\%$; at most $4\%$ elsewhere), the learned map barely at all, and the co-visibility mask changes little. Values are means over training steps $40$k--$50$k; lower is better; \textbf{bold} = best per column.}
\label{tab:cue_ablation}
\begin{adjustbox}{max width=\textwidth}
\begin{tabular}{ll ccc cccccc}
\toprule
 &  & \multicolumn{3}{c}{cue carries} & \multicolumn{4}{c}{LR consistency} & \multicolumn{2}{c}{Disparity} \\
\cmidrule(lr){3-5} \cmidrule(lr){6-9} \cmidrule(lr){10-11}
Variant & channel & sign & expl. & mask & Middleb.-H & ETH3D & KITTI-15 & Booster-H & Middleb.-H & ETH3D \\
 &  &  &  &  & LR-BP-1 & LR-BP-1 & LR-BP-1 & LR-BP-1 & BP-2 & BP-1 \\
\midrule
no cue & -- & -- & -- & -- & 9.36 & 4.99 & 6.68 & 6.77 & \textbf{2.61} & 0.66 \\
\midrule
learned & $\sigma(\mathrm{conv})$ &  &  &  & 9.48 & 4.93 & 6.61 & 6.62 & 2.69 & 0.67 \\
learned + mask & $\sigma(\mathrm{conv}),m$ &  &  & \checkmark & 9.42 & 5.00 & 6.50 & 6.54 & 2.63 & 0.64 \\
unsigned & $|c|$ &  & \checkmark &  & 9.42 & 4.82 & 6.63 & 6.03 & 2.65 & 0.63 \\
unsigned + mask & $|c|,m$ &  & \checkmark & \checkmark & 9.38 & 4.79 & 6.61 & 5.95 & \textbf{2.61} & 0.62 \\
signed & $c$ & \checkmark & \checkmark &  & \textbf{9.06} & 3.29 & 5.53 & 3.65 & 2.71 & 0.59 \\
signed + mask (ours) & $c,m$ & \checkmark & \checkmark & \checkmark & \textbf{9.06} & \textbf{3.24} & \textbf{5.44} & \textbf{3.64} & 2.68 & \textbf{0.58} \\
\bottomrule
\end{tabular}
\end{adjustbox}
\end{table*}

\paragraph{Relation to LRCR's own ablation.} LRCR reports the opposite ordering between subtraction and learned map. Its \emph{non-attention} variant, which ``replaces the learned attention maps with a simple subtraction on the two disparity maps,'' is worse than the learned map on every metric reported at the final recurrent step. The two results are compatible. In LRCR the subtraction is substituted \emph{for} the attention map, so a signed quantity is used where a $[0,1]$ gate is expected and its sign has no natural interpretation, and it is concatenated onto the cost volume, where a negative value competes with matching costs. In a RAFT-style refiner the same quantity enters as an input feature that the update operator can act on, and the sign becomes the useful part.

\clearpage
\section{Consistency Loss Coefficient Ablation}
\label{sec:lambda}

The consistency loss coefficient $\lambda$ (Eq.~\ref{eq:full}) is fixed at $0.5$ everywhere in this paper. This value was chosen through a sweep (see Tab.~\ref{tab:lambda}), changing that one coefficient and nothing else and running the full $50$k schedule at every point. Left-right consistency improves monotonically across the whole range and has not saturated at $\lambda=2$; disparity error does not, and $0.5$ is where the two meet. Below it, consistency degrades steeply (Booster LR-BP-1 $3.64\rightarrow6.78$ at $\lambda=0$) while disparity error on Middlebury, ETH3D and KITTI-15 changes by at most $0.10$; only Booster BP-2 favors a smaller weight ($9.00$ at $\lambda=0$ against $9.29$); above it the trade turns bad, with $\lambda=2$ buying $0.7$ Booster LR-BP-1 and paying $1.3$ Booster BP-2 alongside clear oversmoothing at boundaries and occlusions ($+3.3$ Middlebury edge, $+3.0$ Middlebury occlusion and $+2.8$ Booster occlusion BP-0.5 relative to $0.5$).

\begin{table*}[ht!]
\centering
\caption{\textbf{Sensitivity to the consistency-loss weight $\lambda$.} Every row shares the recipe of Tab.~\ref{tab:ablation}(e) and differs only in $\lambda$ (Eq.~\ref{eq:full}); the $0.5$ row \emph{is} that run. The $0$ row is Tab.~\ref{tab:ablation}(c): the same cross-injection refiner with the consistency loss switched off. \textbf{Finding:} left-right consistency improves monotonically across the full $20\times$ range and has not saturated at $\lambda=2$, but disparity error and the edge/occlusion ROIs turn over: they hold or improve up to $\lambda=0.5$ (except Booster BP-2, which favors smaller weights) and degrade beyond it. The chosen $\lambda=0.5$ is the knee of that trade-off (the largest weight whose consistency gain is essentially free) and is also the minimum of three of the four disparity columns. Values are means over training steps $40$k--$50$k; lower is better; \textbf{bold} = best per column.}
\label{tab:lambda}
\begin{adjustbox}{max width=\textwidth}
\begin{tabular}{l|cccc|cccc|ccc}
\toprule
 & \multicolumn{4}{c|}{Left-Right Consistency} & \multicolumn{4}{c|}{Disparity Error} & \multicolumn{3}{c}{Edge/Occlusion Error} \\
\cmidrule(lr){2-5} \cmidrule(lr){6-9} \cmidrule(lr){10-12}
$\lambda$ & \shortstack{Middleb.-H\\LR-BP-1} & \shortstack{ETH3D\\LR-BP-1} & \shortstack{KITTI-15\\LR-BP-1} & \shortstack{Booster-H\\LR-BP-1} & \shortstack{Middleb.-H\\BP-2} & \shortstack{ETH3D\\BP-1} & \shortstack{KITTI-15\\D1} & \shortstack{Booster-H\\BP-2} & \shortstack{Middleb.-H\\Edg. BP-0.5} & \shortstack{Middleb.-H\\Occ. BP-0.5} & \shortstack{Booster-H\\Occ. BP-0.5} \\
\midrule
$0$\phantom{.00}~~(no loss) & 9.63 & 4.81 & 9.67 & 6.78 & 2.72 & 0.68 & 2.81 & \textbf{9.00} & 32.34 & 33.66 & 51.26 \\
\midrule
$0.1$ & 9.50 & 4.59 & 7.30 & 5.53 & 2.72 & 0.67 & 2.80 & 9.17 & 32.31 & 33.78 & 51.25 \\
$0.25$ & 9.29 & 3.64 & 6.27 & 4.24 & 2.72 & 0.64 & 2.78 & 9.10 & \textbf{32.17} & \textbf{33.30} & \textbf{50.68} \\
$\mathbf{0.5}$~~(ours) & 9.06 & 3.24 & 5.44 & 3.64 & \textbf{2.68} & \textbf{0.58} & \textbf{2.71} & 9.29 & 32.29 & 33.58 & 51.37 \\
$1.0$ & 8.78 & 2.95 & 4.44 & 3.21 & 2.79 & 0.60 & 2.72 & 9.64 & 33.01 & 34.69 & 53.10 \\
$2.0$ & \textbf{8.47} & \textbf{2.60} & \textbf{3.63} & \textbf{2.94} & 3.07 & 0.71 & 2.75 & 10.58 & 35.61 & 36.58 & 54.12 \\
\bottomrule
\end{tabular}
\end{adjustbox}
\end{table*}

\FloatBarrier
\section{Training from Scratch}
\label{sec:scratch}

\begin{figure}[t]
  \centering
  \includegraphics[width=\textwidth]{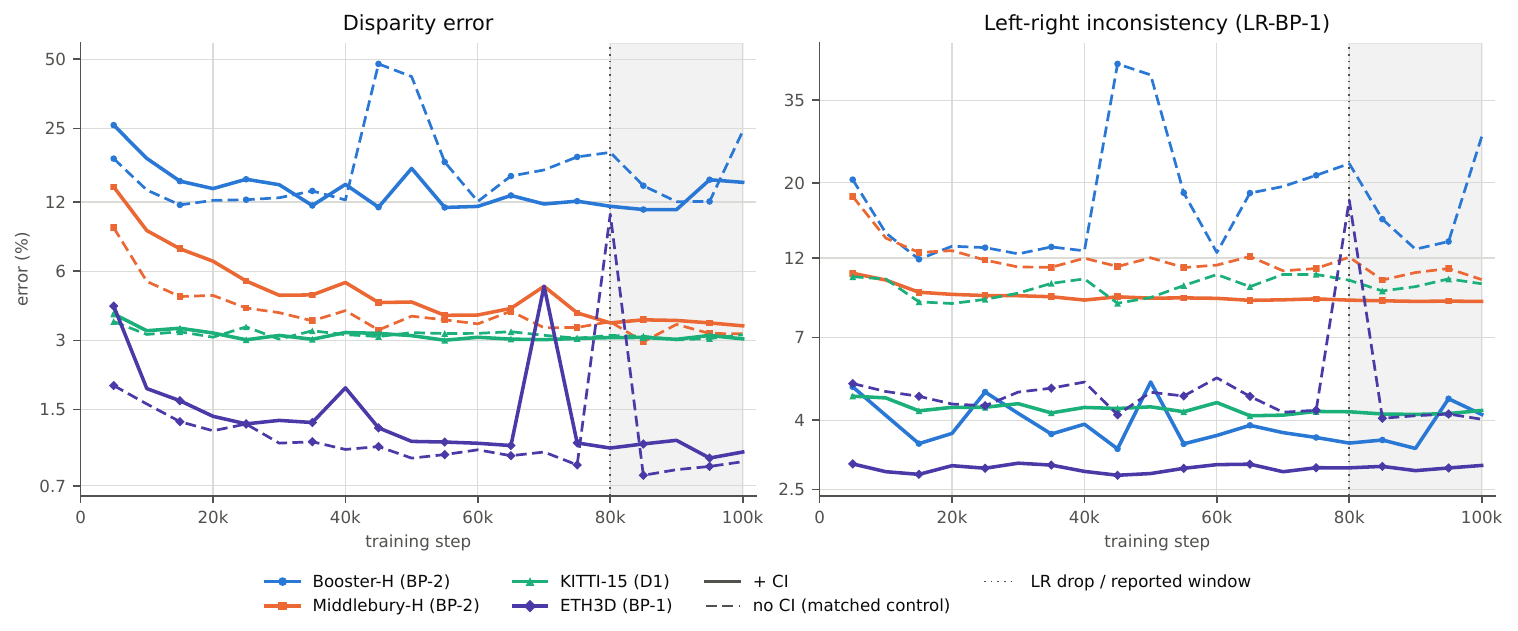}
  \caption{\textbf{Cross-injection trained from scratch.} A matched pair trained end to end for $100$k steps and differing \emph{only} in the cross-injection refiner and the consistency loss, shown on four benchmarks at all $20$ evaluated steps. Color is the benchmark and line style is the arm (\emph{solid} with cross-injection, \emph{dashed} its matched control), so each solid/dashed pair of one color is a controlled comparison. The left panel reports each benchmark's own disparity metric (legend), the right panel LR-BP-1 throughout. The dotted line marks the learning-rate drop at $80$k and the shaded band the final $20$k. \emph{Right:} on every benchmark the solid curve lies below its dashed control at every step. \emph{Left:} cross-injection starts above its control and closes the gap over the run. Vertical axes are logarithmic. These two runs are comparable only to each other.}
  \label{fig:scratch_convergence}
\end{figure}

We view the ability to retrofit cross-injection onto an already-trained matcher as a strength of the method, and it is therefore how we apply it throughout this paper. It is not a requirement, however: Fig.~\ref{fig:scratch_convergence} shows a matched from-scratch pair, demonstrating that the coupling works when the whole network is trained end to end as well. Both arms start from random initialization (except for the frozen, pretrained Depth Anything V2~\cite{yang2024depthanythingv2} monocular branch, as in FoundationStereo), run $100$k steps at an effective batch of $128$ on the FoundationStereo mixture of Sec.~\ref{sec:training_details}, and share every hyperparameter, augmentation and random seed, differing only in the cross-injection refiner and the consistency loss. As expected, convergence in disparity error is slower in early training, which we attribute to the consistency term being unreliable while both views are still badly matched. This could in principle be compensated by trust weighting (downweighting pixels at which the predicted and ground-truth disparities disagree), which we did not explore in this setting. In later training the disparity errors of the two variants largely align, and on Booster, cross-injection is clearly ahead, while left-right consistency is substantially better with cross-injection throughout. Note that $100$k steps is only half of FoundationStereo's original $200$k-step schedule and that we train on a subset of its data composition, so neither run is comparable to the public checkpoint.

\FloatBarrier
\section{Where the Single-View Change Comes From}
\label{sec:regions}

Sec.~\ref{sec:results} reports that cross-injection holds or improves single-view accuracy. This appendix asks where in the image that change occurs. We compare the full method against its matched control (rows~(e) and~(a) of Tab.~\ref{tab:ablation}) over the same six checkpoints from $40$k to $50$k.

We define five image regions of interest, each derived from ground truth and the input image alone. \emph{Occluded} and \emph{out of view} are the two factors of the co-visibility mask of Eq.~\ref{eq:covis} evaluated on ground-truth disparity: a pixel is out of view when its match falls outside the frame, and occluded when it has an in-frame match but fails the single-view uniqueness test. \emph{Depth edges} use the definition of Sec.~\ref{sec:results}, a $2$\,px band around a disparity jump larger than $2$\,px. \emph{Textureless} pixels are those whose mean squared horizontal second luminance difference, over a window of $1\%$ of image width, falls below $2$ on the $0$--$255$ scale. \emph{Remainder} is the complement of the other four. The regions overlap slightly, so this is a decomposition by category rather than a partition, and the shares sum to a little over $100\%$.

Regions are scored at BP-2, computed from separate predictions and pooled over pixels rather than averaged per image, so the all-valid-pixels row differs slightly from Tab.~\ref{tab:ablation} ($-0.08$ against $-0.10$).
Only Middlebury and Booster provide two-sided ground truth, and the decomposition is scored over both views, so they are the only benchmarks it covers.
Each effect is reported beside a noise figure: the same comparison between control checkpoints, which contain no cross-injection. We only draw conclusions where an effect is at least twice its noise.

\paragraph{Cross-injection redistributes error rather than reducing it uniformly.} One effect is clearly larger than the noise on both benchmarks: a mild regression in occluded regions; out-of-view pixels stay within their noise on both. This is likely influenced by the loss: the newly introduced term $\ell_{\mathrm{lrc}}$ (Eq.~\ref{eq:ts}) applies only to co-visible pixels, which therefore receive more signal during backpropagation. A second explanation is that prediction errors may lead to inaccurate co-visibility masks, so that cross-injection is fed residuals computed against falsely matched pixels in the partner view.
Tab.~\ref{tab:ablation}, whose occlusion region is defined differently (a left-right ground-truth check, scored at BP-0.5), shows a change in the same direction that stays within its checkpoint spread ($33.39\rightarrow33.58$ on Middlebury, $51.12\rightarrow51.37$ on Booster). On Middlebury, depth edges and the remainder both improve by more than twice their noise, consistent with CI acting mainly on co-visible pixels. Edges are scored here at BP-2, which targets large errors such as flying pixels, whereas Tab.~\ref{tab:ablation} reports BP-0.5, which targets subtle edge smoothing.
Textureless carries no conclusion: it is at the noise level on Middlebury and below twice it on Booster, and its size differs greatly between the two benchmarks, $72.9\%$ of Booster against $22.4\%$ of Middlebury, because the threshold is absolute.

\begin{table}[ht!]
\centering
\caption{\textbf{Single-view error change by region.} Change in BP-2 (percentage points) for the full method against its matched control (rows~(e) and~(a) of Tab.~\ref{tab:ablation}) as the mean over the same six checkpoints; negative is better. Share is the fraction of valid-ground-truth pixels the region covers, exact over the full evaluation set. Noise is the same comparison between control checkpoints, which contain no cross-injection. \textbf{Bold} marks effects at least twice the noise. The rows overlap and do not sum to the all-valid-pixels row.}
\label{tab:regions}
\footnotesize
\begin{tabular}{l rr r@{\hspace{2em}} rr r}
\toprule
 & \multicolumn{3}{c}{Middlebury-H} & \multicolumn{3}{c}{Booster-H} \\
\cmidrule(lr){2-4} \cmidrule(lr){5-7}
Region & share\,\% & $\Delta$\,BP-2 & noise & share\,\% & $\Delta$\,BP-2 & noise \\
\midrule
all valid pixels & 100.0 & $\mathbf{-0.079}$ & $0.033$ & 100.0 & $+0.065$ & $0.088$ \\
\midrule
textureless & 22.4 & $-0.202$ & $0.202$ & 72.9 & $+0.137$ & $0.073$ \\
depth edges & 9.7 & $\mathbf{-0.045}$ & $0.020$ & 0.9 & $-0.083$ & $0.101$ \\
occluded & 6.7 & $\mathbf{+0.562}$ & $0.013$ & 1.9 & $\mathbf{+0.293}$ & $0.093$ \\
out of view & 4.0 & $-0.170$ & $0.247$ & 5.6 & $-0.194$ & $0.674$ \\
remainder & 62.0 & $\mathbf{-0.064}$ & $0.005$ & 25.1 & $-0.110$ & $0.097$ \\
\bottomrule
\end{tabular}
\end{table}

\begin{table}[ht!]
\centering
\caption{\textbf{Where does cross-injection change the error?} Non-occluded valid left-view pixels, split by whether the control's left and right predictions agree with each other ($|c| \le 1$\,px), for rows (e) and (a) of Tab.~\ref{tab:ablation}, pooled over the six checkpoint pairs. Values are mean absolute disparity error in pixels, for the control and under cross-injection, and their difference $\Delta$ ($+$CI minus control; negative is better). Noise is the same comparison between control checkpoints, which contain no cross-injection. Because the groups are defined by the control's predictions, $\Delta$ is biased by the grouping; the text tests this.}
\label{tab:regions_cells}
\footnotesize
\begin{tabular}{l rrrrr}
\toprule
Pixel group & share\,\% & control & $+$CI & $\Delta$ & noise \\
\midrule
\multicolumn{6}{l}{\textbf{Middlebury-H}} \\
\quad consistent ($|c| \le 1$\,px) & 96.8 & $0.254$ & $0.260$ & $+0.006$ & $0.002$ \\
\quad inconsistent & 3.2 & $5.259$ & $5.036$ & $-0.223$ & $0.017$ \\
\midrule
\multicolumn{6}{l}{\textbf{Booster-H}} \\
\quad consistent ($|c| \le 1$\,px) & 94.7 & $1.319$ & $1.313$ & $-0.006$ & $0.055$ \\
\quad inconsistent & 5.3 & $16.845$ & $14.505$ & $-2.341$ & $1.158$ \\
\bottomrule
\end{tabular}
\end{table}

\paragraph{Does cross-injection correct or entrench pixels that are consistent but wrong?} Left-right consistency and accuracy are not the same property: the two views can agree on the same wrong disparity, and there the difference cross-injection injects is zero, so the cue cannot repair the pixel; and because the consistency term is already satisfied, it can only resist the disparity term's attempt to move either view off the shared value. Sec.~\ref{sec:methods} raises this as the reason a plain symmetric constraint is insufficient; Tab.~\ref{tab:regions_cells} measures it. In practice that resistance is not measurable: the error is unchanged in the consistent group and drops in the inconsistent one. Cross-injection is therefore a targeted correction, not a general accuracy improvement, and since the pixels where the control's two views disagree are only about $3\%$ (Middlebury) and $5\%$ (Booster) of the non-occluded valid left-view pixels, the whole-image change in Tab.~\ref{tab:regions} is correspondingly small. On Middlebury, among pixels where the control's error exceeds $1$\,px, cross-injection lowers the mean error by $4.9\%$ where the control's left and right predictions disagree and by $1.9\%$ where they agree; the next paragraph tests how much of this is selection.

\paragraph{Testing for sampling bias.} Grouping pixels by the control's own predictions biases the comparison in cross-injection's favor: inside a group selected for the control being wrong, part of that error is random, and cross-injection does not repeat it. Repeating the analysis with the groups defined by cross-injection's predictions instead turns the bias in the control's favor. For pixels that are consistent but wrong, the sign then flips (Middlebury $-1.9\%\rightarrow+1.4\%$, Booster $-1.4\%\rightarrow+3.3\%$), so we find no reliable effect there: cross-injection neither repairs nor entrenches these pixels. On inconsistent pixels, cross-injection remains at least as accurate as the control even with the bias against it (Middlebury $2.8\%$, Booster $0.3\%$ lower mean error), so the improvement there is not a selection artifact; its size, however, depends on which arm defines the groups, most strongly on Booster.

\end{appendices}

\end{document}